\pdfoutput=1
\documentclass[11pt]{article}

\usepackage[preprint]{acl}
\usepackage{times}
\usepackage{latexsym}
\usepackage[T1]{fontenc}
\usepackage[utf8]{inputenc}
\usepackage{inconsolata}
\usepackage{microtype}
\usepackage{graphicx}
\usepackage{booktabs}
\usepackage{multirow}
\usepackage{hyperref}
\usepackage{amsmath}
\usepackage{amssymb}
\usepackage{mathtools}
\usepackage{amsthm}
\usepackage{algorithm}
\usepackage{algpseudocode}
\usepackage[capitalize,noabbrev]{cleveref}
\usepackage{xcolor}
\usepackage{tcolorbox}
\tcbuselibrary{skins,breakable}
\usepackage{colortbl}
\usepackage{caption}
\usepackage{enumitem}
\usepackage{listings}
\crefname{lstlisting}{Listing}{Listings}
\lstdefinestyle{pseudo}{language=Python, basicstyle=\ttfamily\scriptsize, keywordstyle=\color{blue!60!black}\bfseries, commentstyle=\color{green!40!black}\itshape, frame=single, rulecolor=\color{gray!50}, backgroundcolor=\color{gray!4}, breaklines=true, columns=fullflexible, xleftmargin=0.5em, xrightmargin=0.5em}
\lstdefinestyle{guide}{basicstyle=\ttfamily\scriptsize, breakindent=1.5em, breakatwhitespace=true, frame=single, rulecolor=\color{gray!50}, backgroundcolor=\color{blue!2}, breaklines=true, columns=fullflexible, xleftmargin=0.5em, xrightmargin=0.5em}
\usepackage{placeins}

\IfFileExists{latexml.sty}{\usepackage{latexml}}{\newif\iflatexml}
\iflatexml
  \renewcommand{\resizebox}[3]{#3}
  \renewcommand{\multirow}[3]{#3}
  \renewcommand{\rotatebox}[3][]{#3}
  \newenvironment{skillbox}[2][]{\par\smallskip\noindent\textbf{#2}\par\renewcommand{\hfill}{\quad}}{\par\smallskip}
\else
  \newenvironment{skillbox}[2][]{\begin{tcolorbox}[#1,title={#2}]}{\end{tcolorbox}}
\fi

\DeclareMathOperator*{\argmax}{arg\,max}

\newcommand{\vw}{\mathbf{w}}
\newcommand{\vm}{\mathbf{m}}
\newcommand{\cP}{\mathcal{P}}
\newcommand{\cD}{\mathcal{D}}
\newcommand{\hv}{\mathrm{HV}}
\newcommand{\hvc}{\mathrm{HVC}}
\colorlet{stuck}{red!9}
\colorlet{meanrow}{gray!14}
\theoremstyle{plain}
\newtheorem{proposition}{Proposition}
\definecolor{ablbase}{HTML}{CFEB85}
\definecolor{ablplus}{HTML}{AFDD70}
\definecolor{ablminus}{HTML}{FFF6B0}

\title{MOCHA: Multi-Objective Chebyshev Annealing \\ for Agent Skill Optimization}

\iflatexml
\author{
    Md Mehrab Tanjim, Jayakumar Subramanian, Branislav Kveton, Xiang Chen, Subhojyoti Mukherjee, Anlan Zhang, Sungchul Kim, Sunav Choudhury, Somdeb Sarkhel \\
    Adobe Research \\
    \texttt{\{tanjim, xiangche, jasubram, subhomuk, kveton, anlanz, sukim, sarkhel, schoudha\}@adobe.com}
}
\else
\author{
    Md Mehrab Tanjim, Jayakumar Subramanian, Branislav Kveton \\
    \textbf{Xiang Chen}, \textbf{Subhojyoti Mukherjee}, \textbf{Anlan Zhang} \\
    \textbf{Sungchul Kim}, \textbf{Sunav Choudhury}, \textbf{Somdeb Sarkhel} \\
    Adobe Research \\
    \texttt{\{tanjim, jasubram, kveton, xiangche, subhomuk, } \\
    \texttt{anlanz, sukim,  schoudha, sarkhel\}@adobe.com}
}
\fi

\begin{document}
\maketitle

\begin{figure*}[t]
    \centering
    \iflatexml
    \includegraphics[width=0.63\textwidth]{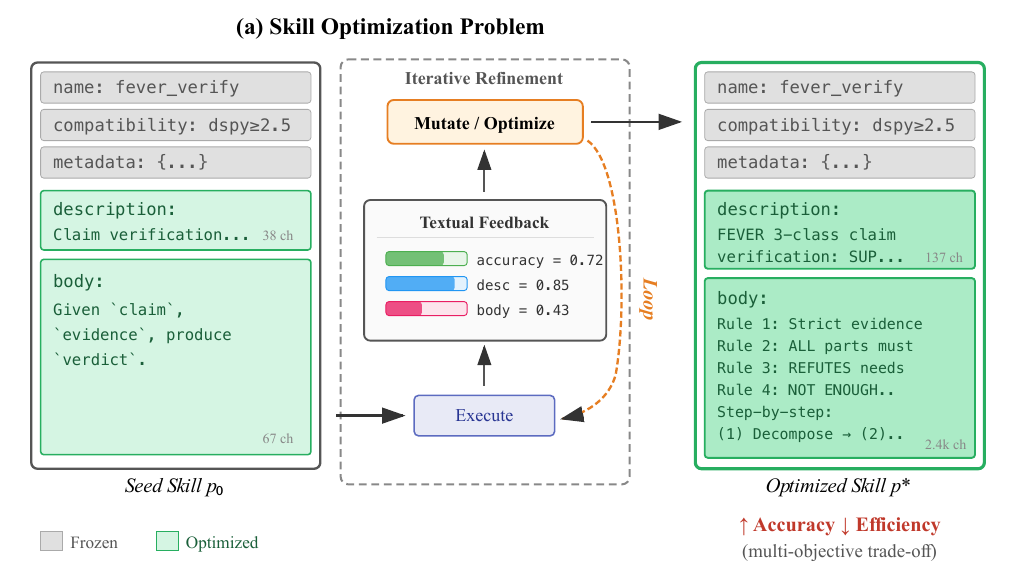}\quad
    \includegraphics[width=0.35\textwidth]{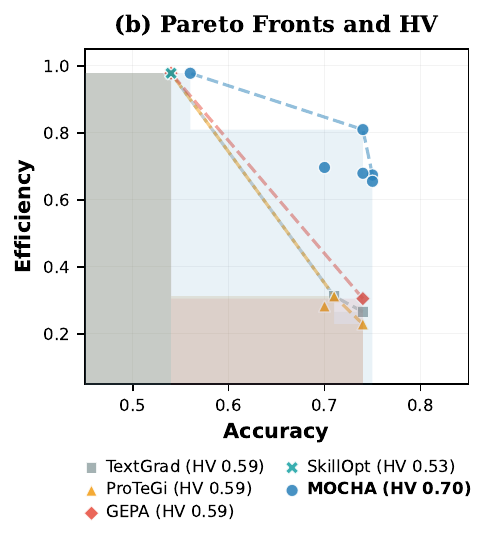}
    \else
    \begin{minipage}[c]{0.63\textwidth}
        \includegraphics[width=\linewidth]{figures/fig_skill_problem_a.pdf}
    \end{minipage}\hfill
    \begin{minipage}[c]{0.35\textwidth}
        \includegraphics[width=\linewidth]{figures/teaser_hv.pdf}
    \end{minipage}
    \fi
    \caption{\textbf{(a)}~Skill optimization rewrites a skill from execution feedback; the optimized skill gains accuracy but spends more of its length budget, a multi-objective trade-off. \textbf{(b)}~Accuracy--efficiency Pareto fronts and their hypervolume (HV). Each point is an optimized skill; shading marks the region each method's front dominates. The baselines trade efficiency for accuracy along a single path; MOCHA spans the trade-off and raises HV.}
    \label{fig:teaser}
\end{figure*}

\begin{abstract}
LLM agents organize behavior through \emph{skills}---structured natural-language specifications that govern how an agent reasons, retrieves, and responds. Unlike monolithic prompts, skills are multi-field artifacts with platform limits: descriptions are indexed for routing, instruction bodies are loaded into a shared context window, and every extra line costs the agent context it could spend elsewhere. Skill optimization is therefore \emph{inherently} multi-objective: a skill should be accurate and economical at once. Existing prompt optimizers collapse these goals into a single score, and we find that this makes them stall: in most of their runs, \textbf{1000 rollouts yield zero progress} over the seed skill. We introduce \textbf{MOCHA} (\textbf{M}ulti-\textbf{O}bjective \textbf{CH}ebyshev \textbf{A}nnealing), which selects skills to improve by Chebyshev scalarization and accepts improvements that expand the Pareto front, annealing from exploration to exploitation. Chebyshev scalarization scores a skill by its largest weighted shortfall from a perfect skill; with random weights, it can select any skill on the Pareto front for improvement, including trade-offs that an average would never favor.
Across six diverse skills and two models, MOCHA escapes the seed in nearly every run, achieving \textbf{7.9\%} higher mean accuracy than the strongest baseline (up to \textbf{23.7\%} on DebugBench with GPT 5.5 and \textbf{14.9\%} on FEVER with Claude Haiku~4.5) while discovering Pareto fronts with the highest hypervolume and nearly \textbf{twice} as many Pareto-optimal skills.
\end{abstract}

\section{Introduction}

The dominant abstraction in early LLM applications was the \emph{prompt}: a monolithic string optimized for a single task~\citep{brown2020language, wei2022chain}. As agents have grown more capable, a richer abstraction has emerged: the \emph{skill}~\citep{wang2023voyager, xia2026skillrl, anthropic2025claude}. A skill pairs a \emph{description}, which the agent uses to decide when the skill applies, with an instruction \emph{body} that tells it how to act. Modern agent frameworks organize their entire repertoire this way: a coding agent chooses among debugging, refactoring, and explanation skills; a support agent routes between policy lookup and escalation.

Because skills are written in natural language, prompt optimization~\citep{zhou2023large, pryzant2023automatic, khattab2024dspy, opsahl2024optimizing} can refine them. But skills are not single-objective artifacts. Descriptions are truncated in routing indexes, bodies are capped in length, and skills that load together share one context window, so a verbose skill takes room from its neighbors~\citep{anthropic2025claude}. Longer instructions often raise accuracy; they also spend the budget a skill is meant to respect. \emph{Every author of a deployed skill faces this tension}, and the useful output of an optimizer is therefore not one skill but a \emph{Pareto front} of skills, from which a practitioner picks the trade-off that suits a deployment. The natural measure of such a front is its \emph{hypervolume} (HV), the region of objective space it covers (\Cref{fig:teaser}b).

Reflection-based optimizers~\citep{pryzant2023automatic, yuksekgonul2024textgrad, agrawal2025gepa, yang2026skillopt} are the natural tools for this problem: they revise text from execution feedback and need no access to model weights. One can even give them feedback on every objective---our experiments do exactly that---yet this alone is not enough. All of them ultimately judge a rewrite by a single score, and we find that this choice makes them stall. TextGrad, ProTeGi, and GEPA return the seed skill unchanged in 112 of their 180 runs, and SkillOpt in 50 of 55: 1000 rollouts\footnote{Following GEPA~\citep{agrawal2025gepa}, a \emph{rollout} is one metric call: running a skill on one example and scoring the result, whether the example belongs to a training minibatch or to the validation set. The optimization budget is the total number of rollouts, so ``budget'' and ``rollouts'' are interchangeable.} of optimization yield zero progress. The reason is simple. A useful rewrite adds instructions, so it gains accuracy but loses efficiency; when both are folded into one score, the loss usually outweighs the gain, and the rewrite is rejected. The seed, short and reasonably accurate, becomes a trap.

\textbf{Key insight: skill optimization needs selection that understands trade-offs.} A rewrite that is more accurate but longer is not worse; it is a different point on the Pareto front.
MOCHA keeps every such point. At each step it chooses a skill to improve by Chebyshev scalarization with a random weighting, which can reach any skill on the Pareto front~\citep{miettinen1999nonlinear}. Early on it \emph{explores}, accepting any rewrite that expands the Pareto front; as the budget is spent it \emph{exploits}, accepting only rewrites that improve the skill under the weighting that selected it.

\textbf{Contributions.} This paper makes three contributions:
\begin{itemize}[nosep,leftmargin=1.2em]
    \item \textbf{Problem and diagnosis.} We cast skill optimization as a multi-objective problem over accuracy and token efficiency, and show why optimizers that fold these into one score stall at the seed.
    \item \textbf{Method.} MOCHA brings Chebyshev scalarization and hypervolume-based acceptance, well studied in continuous optimization~\citep{miettinen1999nonlinear, lin2024smooth, mukherjee2024multi}, to the discrete and expensive setting of natural-language skills.
    \item \textbf{Evaluation.} Across six diverse skills and two model families, MOCHA escapes the seed in 58 of 60 runs, achieves 7.9\% higher mean accuracy than the strongest baseline (up to 23.7\% on DebugBench with GPT 5.5 and 14.9\% on FEVER with Claude Haiku~4.5), and finds Pareto fronts with the highest hypervolume and nearly twice as many Pareto-optimal skills.
\end{itemize}

\section{Related Work}
\label{sec:related_work}

\textbf{Prompt optimization.}
Prompt optimizers either propose and rank candidates~\citep{zhou2023large, yang2024large, guo2024connecting, opsahl2024optimizing}, learn from rewards~\citep{deng2022rlprompt}, or refine text through reflection on execution feedback~\citep{pryzant2023automatic, yuksekgonul2024textgrad, cheng2024trace, agrawal2025gepa}. MOCHA belongs to the last family. Its members differ mainly in which candidate they refine: ProTeGi uses UCB beam search, TextGrad refines the current best, and GEPA samples from a Pareto front defined over \emph{validation examples, not metrics}. All of them rank candidates by a single score that averages the metrics, so a rewrite that trades one metric for another is judged only by its net change and the trade-off itself is lost; MOCHA ranks candidates over the \emph{objectives} jointly.

\textbf{Agent skills.}
An agent can learn from feedback by updating its model's weights or by updating the skills that govern its behavior~\citep{xia2026skillrl, wang2023voyager}. Skill discovery methods such as SkillRL~\citep{xia2026skillrl}, Voyager~\citep{wang2023voyager}, and EUREKA~\citep{ma2023eureka} need many trajectories to extract a single reusable skill, and approaches that fine-tune the model are unavailable when the agent is a closed API model. Refining the text of existing skills is therefore the practical lever. SkillOpt~\citep{yang2026skillopt} does exactly this through small edits, but, like the prompt optimizers above, it collapses its objectives into their average, and it faces an even higher bar: an edit must raise that average on the full validation set rather than on a small minibatch. MOCHA also refines through API access alone, but judges them on all objectives jointly and returns a front of trade-offs.

\textbf{Multi-objective optimization.}
Evolutionary methods approximate Pareto fronts in multi-objective optimization by non-dominated sorting (NSGA-II;~\citealp{deb2002fast}), by decomposing the problem into weighted subproblems (MOEA/D;~\citealp{zhang2007moead}), or by selecting for hypervolume (SMS-EMOA;~\citealp{emmerich05emo, beume2007sms}).
Evolutionary methods, however, evolve large populations over many generations, which is affordable only when evaluations are cheap. Here each candidate skill costs dozens of LLM calls to evaluate, and a whole run affords only a few dozen candidates. MOCHA therefore keeps the two key ideas, decomposition for selection and hypervolume for acceptance to effectively push the pareto front within budget (\Cref{sec:method}).

\section{Method}
\label{sec:method}

\textbf{Notation.} Let $p \in \mathcal{S}$ denote a skill definition, $\mathcal{S}$ the set of all skill definitions, and $m_1(p), \ldots, m_M(p) \in [0,1]$ the values of $p$ on $M$ metrics, each to be maximized; we write $\vm(p) = (m_1(p), \ldots, m_M(p))$. \Cref{fig:teaser}a and \Cref{fig:skill_compare} show example skills, and \Cref{app:qualitative} shows one for every task.

\subsection{\texorpdfstring{Skill Optimization as Multi-Objective Optimization}{Skill Optimization as Multi-Objective Optimization}}
\label{sec:problem}

A skill consists of text fields, such as a description used to decide when it applies and a body of instructions, and the optimizer may rewrite any of them. A skill $p'$ \emph{dominates} $p$, written $\vm(p') \succ \vm(p)$, if $m_j(p') \ge m_j(p)$ for every $j$ and $m_j(p') > m_j(p)$ for some $j$~\citep{emmerich2018tutorial}. We seek the Pareto-optimal skills,
\begin{equation}
    \mathcal{S}^* = \{p \in \mathcal{S} : \nexists\, p' \in \mathcal{S} \text{ with } \vm(p') \succ \vm(p)\}.
\label{eq:pareto}
\end{equation}
Finding $\mathcal{S}^*$ is hard: $\mathcal{S}$ contains every possible text and is effectively infinite, the metrics are black boxes without gradients, and scoring a single skill takes many LLM calls.
Rather than committing to a single optimum, we take the \emph{a-posteriori} view of multi-objective optimization~\citep{miettinen1999nonlinear}: the optimizer returns the non-dominated skills it finds, and the practitioner chooses the trade-off that suits a deployment. The formulation is generic; any scores in $[0,1]$ can serve as metrics. Our experiments use task accuracy and the token efficiency of the skill's two fields (\Cref{sec:setup}).

We measure a set of skills $\cP$ by its \emph{hypervolume} (HV), the volume of objective space it dominates with respect to the origin, and a new skill by its \emph{hypervolume contribution} (HVC), the volume it adds:
\begin{align}
    \hv(\cP) &= \lambda\Big(\textstyle\bigcup_{p \in \cP} \prod_{j=1}^{M} [0, m_j(p)]\Big), \label{eq:hv} \\
    \hvc(p', \cP) &= \hv(\cP \cup \{p'\}) - \hv(\cP), \label{eq:hvc}
\end{align}
where $\lambda$ is the Lebesgue measure. Hypervolume is the only known single-number measure of a Pareto front that strictly increases whenever the front improves in the Pareto sense~\citep{zitzler2003performance}, and $\hvc(p', \cP) > 0$ exactly when no member of $\cP$ dominates $p'$. For $M \le 3$, both can be computed exactly in polynomial time~\citep{guerreiro2021hypervolume}.

\textbf{Estimating the metrics.}
We use three disjoint splits: a small training \emph{minibatch} screens each candidate skill cheaply (\Cref{sec:expansion}); a \emph{validation} set gives every kept skill its score vector $\vm(p)$, which defines the Pareto front and guides the search; and a \emph{test} set is used only for reporting (\Cref{sec:setup}).

\subsection{\texorpdfstring{Algorithm}{Algorithm}}
\label{sec:overview}

MOCHA (\Cref{alg:main}) keeps a pool $\cP$ of skills, starting from the seed $p_0$, and repeats three steps until the budget is spent. It chooses a skill in the pool to improve, the \emph{parent} (\Cref{sec:parent}); it rewrites the parent into a \emph{child} (\Cref{sec:child}); and it decides whether the child joins the pool (\Cref{sec:expansion}). Each step answers one design question: which skill to improve, how to improve it, and which improvements to keep. MOCHA returns the pool's Pareto front; the practitioner picks a skill by their own priorities, e.g., by maximizing the Chebyshev score under their weights. Throughout, ``Pareto front'' means the front of the skills found.

\begin{algorithm}[t]
\caption{MOCHA}
\label{alg:main}
\small
\begin{algorithmic}[1]
\Require seed skill $p_0$, budget $B$
\State $\cP \gets \{p_0\}$; score $p_0$ on validation
\While{budget remains}
    \State draw $\vw$; \ $p \gets \argmax_{q \in \cP} s_\vw(q)$ \Comment{parent}
    \State $p' \gets$ \textsc{Rewrite}$(p)$ from minibatch feedback \Comment{child}
    \If{$\alpha(b) > \epsilon$} \Comment{explore; $b$: budget spent}
        \State keep $p'$ if $\hvc(p', \{p\}) > \alpha(b)$
    \Else \Comment{exploit}
        \State keep $p'$ if $s_\vw(p') {>} s_\vw(p)$
    \EndIf
    \State if kept: add $p'$ to $\cP$ with validation scores $\vm(p')$
\EndWhile
\State \Return Pareto front of $\cP$
\end{algorithmic}
\end{algorithm}

\subsection{\texorpdfstring{Parent Selection}{Parent Selection}}
\label{sec:parent}

MOCHA draws a weighting $\vw$ uniformly from the simplex, $\vw \sim \mathrm{Dirichlet}(\mathbf{1})$, and picks as parent the pool member with the largest \emph{Chebyshev score}, computed from its validation scores $\vm(p)$,
\begin{equation}
    s_\vw(p) = 1 - \max_{j} \, w_j \, \big(1 - m_j(p)\big),
\label{eq:chebyshev}
\end{equation}
one minus the largest weighted gap between $p$ and a perfect skill, whose metrics all equal 1. The parent is thus the skill whose largest weighted gap is the smallest, and excelling on one objective cannot make up for neglecting another.

\textbf{Why a random weighting?} Since $\vw$ is a vector, it defines a \emph{direction} in objective space, the relative importance of the objectives, and $s_\vw$ favors the skill that is best in that direction. A fixed $\vw$ would always choose the same parent and improve the Pareto front in one direction only. Drawing $\vw$ at random varies the direction from step to step, so that every skill on the Pareto front can be chosen (\Cref{prop:cheb}) and the search covers the full range of trade-offs. Sampling $\vw$ uniformly needs no tuning and treats the objectives symmetrically.

\begin{proposition}[Chebyshev completeness~\citep{miettinen1999nonlinear}]
\label{prop:cheb}
For any pool $\cP$ and any $p^*$ on its Pareto front, there is a weighting $\vw$ under which $p^*$ maximizes $s_\vw$ over $\cP$.
\end{proposition}

A weighted sum $\sum_j w_j m_j$, by contrast, cannot select Pareto-optimal points in non-convex regions of the front under any weighting. \Cref{prop:cheb} guarantees \emph{selection}, not reachability: every non-dominated skill in the pool can be chosen for further improvement. Whether rewriting discovers a given Pareto-optimal skill is an empirical matter, which the exploration rule (\Cref{sec:expansion}) addresses by widening the pool.

\subsection{\texorpdfstring{Child Generation}{Child Generation}}
\label{sec:child}

The child $p'$ is obtained by rewriting the parent. MOCHA treats the rewrite as a black box. In our experiments, an LLM receives the parent, per-example feedback on a minibatch of $n$ training examples, and each field's length against its limit, and returns a revised skill. Because the parent lies on the current Pareto front, a rewrite that fixes its observed failures can improve the front. All compared methods share this operator, so differences come from selection and acceptance alone.

\begin{figure}[t]
    \centering
    \includegraphics[width=0.72\linewidth]{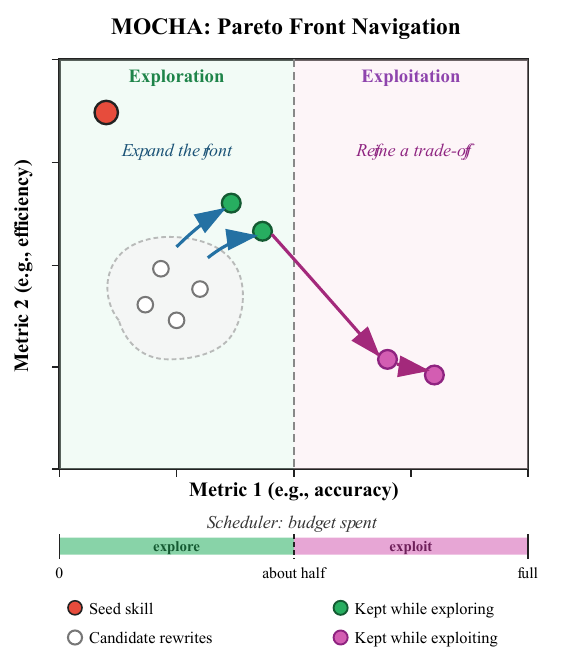}
    \caption{\textbf{MOCHA's two phases.} A budget-driven scheduler anneals from exploration to exploitation. During roughly the first half of the budget, MOCHA keeps rewrites that expand the Pareto front in any direction (green); afterwards, it keeps only rewrites that improve the trade-off they were chosen for (purple).}
    \label{fig:schedule}
\end{figure}
\subsection{\texorpdfstring{Pool Expansion}{Pool Expansion}}
\label{sec:expansion}

A child joins the pool only if it improves on its parent. Both acceptance rules below are applied to the parent's and the child's scores on the minibatch, a cheap screen. Only a kept child is scored on the validation set and added to $\cP$. An iteration thus costs $2n$ rollouts, plus $|\cD_{\mathrm{val}}|$ when a child is kept. MOCHA uses one of two rules:
\begin{itemize}[nosep,leftmargin=1.2em]
    \item \emph{Explore}: keep $p'$ if $\hvc(p', \{p\}) > \alpha(b)$ (\Cref{eq:hvc}), i.e., if it adds enough hypervolume over its parent by improving any objective, regardless of the drawn $\vw$; $\alpha(b)$ is a threshold annealed as described below.
    \item \emph{Exploit}: keep $p'$ if $s_\vw(p') > s_\vw(p)$ (\Cref{eq:chebyshev}), i.e., if it raises the same Chebyshev score that selected $p$, improving the parent in the direction $\vw$.
\end{itemize}

\textbf{Why two rules?} They serve different stages of the search, and we find empirically that each contributes: hypervolume alone yields larger Pareto fronts of less accurate skills, and Chebyshev alone more accurate skills on smaller fronts (\Cref{sec:ablation}). Exploration comes first. The Pareto front starts as a single point, and with a limited budget only a few weightings are drawn, so a rule tied to $\vw$ would reject children that improve the parent in a direction not drawn, leaving parts of the front unimproved. Hypervolume rewards growth in any direction without depending on the drawn $\vw$. Exploitation follows: once the Pareto front has several points, Chebyshev selection and acceptance together improve each of them in its own direction, refining the trade-offs already found.

\textbf{Why annealing?} Almost any rewrite of the seed expands the initial Pareto front: it gains accuracy and loses efficiency, so its parent does not dominate it (\Cref{sec:why}). Yet each kept child costs a validation pass, a tenth of our budget. Keeping every such child would spend the budget on marginal skills, so MOCHA anneals the threshold $\alpha(b)$: it starts high, so that only children adding substantial hypervolume are kept, and decays toward zero as the budget $b$ is spent. Once $\alpha(b)$ falls below a small $\epsilon$, after about half of the budget, MOCHA switches to exploitation (\Cref{fig:schedule}; details in \Cref{app:implementation}).

\section{Experiments}
\label{sec:experiments}

\subsection{Setup}
\label{sec:setup}

\textbf{Skills.}
Each skill follows the SKILL.md format~\citep{anthropic2025claude}: YAML metadata, a description, and a Markdown body. We optimize the description and body of six skills: graduate science QA (GPQA;~\citealp{rein2023gpqa}), theorem-based reasoning (TheoremQA;~\citealp{chen2023theoremqa}), claim verification (HoVer;~\citealp{jiang2020hover}; FEVER;~\citealp{thorne2018fever}), multi-hop QA (HotpotQA;~\citealp{yang2018hotpotqa}), and code repair (DebugBench;~\citealp{tian2024debugbench}). Each run samples 100 training, 100 validation, and 100 test examples.

\textbf{Objectives.}
We optimize $M{=}3$ objectives: \emph{accuracy} on the task (F1 for HotpotQA and pass rate for DebugBench), and the \emph{token efficiency} of the description and of the body, $\max(0, 1 - \ell/L)$, where $\ell$ is the field's length in characters and $L$ its platform limit (1{,}024 and 5{,}000).

\textbf{Models and budget.}
Skills are executed by two models from different LLM families to show generalization: Claude Haiku~4.5, and GPT 5.5. Every method has the same budget of 1000 rollouts, i.e., metric calls on minibatch and validation examples alike (\citealp{agrawal2025gepa}; see the accounting in \Cref{sec:method}), and is run with five seeds. Minibatches have 10 examples, and MOCHA uses one configuration for every task and model (\Cref{app:implementation}).

\textbf{Reporting.}
For accuracy, each method deploys the first kept skill with the best \emph{validation} accuracy, and we report its test accuracy; no test information is used for selection. This single-skill report allows a direct comparison with single-objective baselines; a practitioner can instead pick any skill on the front, and where the baselines stall their front is the seed alone. What each method found as a whole is measured by the \textbf{hypervolume} (HV) of all its kept skills over the three objectives and by the \textbf{Pareto-front size} (\#PF), the number of those skills that are non-dominated, both on the test set.

\textbf{Baselines.}
We compare with \textbf{TextGrad}~\citep{yuksekgonul2024textgrad}, \textbf{ProTeGi}~\citep{pryzant2023automatic}, \textbf{GEPA}~\citep{agrawal2025gepa}, and \textbf{SkillOpt}~\citep{yang2026skillopt}, a recent skill optimizer.
They differ from MOCHA mainly in how they select and accept candidates, all of which they score by the average of the three objectives: TextGrad and SkillOpt select greedily (TextGrad on the minibatch, SkillOpt on the validation set), ProTeGi uses UCB beam search, and GEPA keeps a Pareto front over validation \emph{examples} rather than over the objectives themselves.

\begin{table*}[t]
\centering
\caption{\textbf{Main results} over six skills and two models (mean$\pm$std over five seeds). Each block reports one metric (higher is better) for every skill, followed by the mean (grey). \colorbox{stuck}{Red}: the method returned the seed skill in every run. \textbf{Bold}: best per row.}
\label{tab:main}
\footnotesize
\setlength{\tabcolsep}{2.4pt}
\resizebox{\textwidth}{!}{\begin{tabular}{@{}l cccccc c cccccc c@{}}
\toprule
 & \multicolumn{6}{c}{\textbf{Claude Haiku~4.5}} & & \multicolumn{6}{c}{\textbf{GPT-5.5}} & \\
\cmidrule(lr){2-7}\cmidrule(lr){9-14}
\textbf{Skill} & \textbf{Seed} & \textbf{TextGrad} & \textbf{ProTeGi} & \textbf{GEPA} & \textbf{SkillOpt} & \textbf{MOCHA} & & \textbf{Seed} & \textbf{TextGrad} & \textbf{ProTeGi} & \textbf{GEPA} & \textbf{SkillOpt} & \textbf{MOCHA} & \\
\midrule
GPQA & .521\scriptsize$\pm$.022 & \cellcolor{stuck}.521\scriptsize$\pm$.022 & \cellcolor{stuck}.521\scriptsize$\pm$.022 & \cellcolor{stuck}.521\scriptsize$\pm$.022 & \cellcolor{stuck}.521\scriptsize$\pm$.022 & \textbf{.536}\scriptsize$\pm$.024 &  & .811\scriptsize$\pm$.014 & \textbf{.812}\scriptsize$\pm$.013 & .811\scriptsize$\pm$.014 & .811\scriptsize$\pm$.014 & \cellcolor{stuck}.811\scriptsize$\pm$.014 & .811\scriptsize$\pm$.014 & \\
TheoremQA & .471\scriptsize$\pm$.033 & .506\scriptsize$\pm$.033 & .507\scriptsize$\pm$.040 & .497\scriptsize$\pm$.049 & .509\scriptsize$\pm$.107 & \textbf{.526}\scriptsize$\pm$.030 &  & .570\scriptsize$\pm$.023 & .599\scriptsize$\pm$.054 & .604\scriptsize$\pm$.047 & .646\scriptsize$\pm$.064 & \cellcolor{stuck}.570\scriptsize$\pm$.023 & \textbf{.694}\scriptsize$\pm$.027 & \\
HoVer & .559\scriptsize$\pm$.013 & \cellcolor{stuck}.559\scriptsize$\pm$.013 & \cellcolor{stuck}.559\scriptsize$\pm$.013 & \cellcolor{stuck}.559\scriptsize$\pm$.013 & \cellcolor{stuck}.559\scriptsize$\pm$.013 & \textbf{.571}\scriptsize$\pm$.008 &  & .595\scriptsize$\pm$.010 & \cellcolor{stuck}.595\scriptsize$\pm$.010 & .601\scriptsize$\pm$.022$^{4}$ & .606\scriptsize$\pm$.019$^{4}$ & \cellcolor{stuck}.595\scriptsize$\pm$.010 & \textbf{.644}\scriptsize$\pm$.055 & \\
HotpotQA & .303\scriptsize$\pm$.034 & .395\scriptsize$\pm$.036 & .390\scriptsize$\pm$.036 & \textbf{.396}\scriptsize$\pm$.026 & \cellcolor{stuck}.303\scriptsize$\pm$.034 & .389\scriptsize$\pm$.030 &  & .509\scriptsize$\pm$.048 & \textbf{.600}\scriptsize$\pm$.045 & .570\scriptsize$\pm$.036 & .560\scriptsize$\pm$.039 & .587\scriptsize$\pm$.031 & .594\scriptsize$\pm$.040 & \\
FEVER & .571\scriptsize$\pm$.029 & \cellcolor{stuck}.571\scriptsize$\pm$.029 & \cellcolor{stuck}.571\scriptsize$\pm$.029 & \cellcolor{stuck}.571\scriptsize$\pm$.029 & \cellcolor{stuck}.571\scriptsize$\pm$.029 & \textbf{.604}\scriptsize$\pm$.015 &  & .648\scriptsize$\pm$.028 & \textbf{.687}\scriptsize$\pm$.029 & .669\scriptsize$\pm$.044 & .675\scriptsize$\pm$.048 & \cellcolor{stuck}.648\scriptsize$\pm$.028 & .684\scriptsize$\pm$.033 & \\
DebugBench & .538\scriptsize$\pm$.006 & \cellcolor{stuck}.538\scriptsize$\pm$.006 & \cellcolor{stuck}.538\scriptsize$\pm$.006 & \cellcolor{stuck}.538\scriptsize$\pm$.006 & \cellcolor{stuck}.538\scriptsize$\pm$.006 & \textbf{.560}\scriptsize$\pm$.007 &  & .385\scriptsize$\pm$.017 & .416\scriptsize$\pm$.049 & .440\scriptsize$\pm$.066 & .452\scriptsize$\pm$.069 & \cellcolor{stuck}.385\scriptsize$\pm$.017 & \textbf{.530}\scriptsize$\pm$.048 & \\
\rowcolor{meanrow}\textit{Mean} & .494 & .515 & .514 & .514 & .500 & \textbf{.531} &  & .586 & .618 & .616 & .625 & .599 & \textbf{.660} & \multirow{-7}{*}{\rotatebox[origin=c]{90}{\textbf{HV}}} \\
\midrule
GPQA & .592\scriptsize$\pm$.025 & \cellcolor{stuck}.592\scriptsize$\pm$.025 & \cellcolor{stuck}.592\scriptsize$\pm$.025 & \cellcolor{stuck}.592\scriptsize$\pm$.025 & \cellcolor{stuck}.592\scriptsize$\pm$.025 & \textbf{.636}\scriptsize$\pm$.056 &  & .860\scriptsize$\pm$.014 & \textbf{.866}\scriptsize$\pm$.015 & .858\scriptsize$\pm$.013 & .858\scriptsize$\pm$.013 & \cellcolor{stuck}.860\scriptsize$\pm$.014 & .852\scriptsize$\pm$.032 & \\
TheoremQA & .534\scriptsize$\pm$.037 & .672\scriptsize$\pm$.087 & .690\scriptsize$\pm$.105 & .656\scriptsize$\pm$.129 & .626\scriptsize$\pm$.085 & \textbf{.762}\scriptsize$\pm$.046 &  & .616\scriptsize$\pm$.039 & .642\scriptsize$\pm$.063 & .700\scriptsize$\pm$.086 & .762\scriptsize$\pm$.092 & \cellcolor{stuck}.616\scriptsize$\pm$.039 & \textbf{.808}\scriptsize$\pm$.035 & \\
HoVer & .618\scriptsize$\pm$.015 & \cellcolor{stuck}.618\scriptsize$\pm$.015 & \cellcolor{stuck}.618\scriptsize$\pm$.015 & \cellcolor{stuck}.618\scriptsize$\pm$.015 & \cellcolor{stuck}.618\scriptsize$\pm$.015 & \textbf{.660}\scriptsize$\pm$.043 &  & .624\scriptsize$\pm$.008 & .626\scriptsize$\pm$.010 & .626\scriptsize$\pm$.024 & .648\scriptsize$\pm$.057 & \cellcolor{stuck}.624\scriptsize$\pm$.008 & \textbf{.696}\scriptsize$\pm$.080 & \\
HotpotQA & .336\scriptsize$\pm$.038 & .592\scriptsize$\pm$.030 & \textbf{.622}\scriptsize$\pm$.041 & .602\scriptsize$\pm$.029 & \cellcolor{stuck}.336\scriptsize$\pm$.038 & .600\scriptsize$\pm$.034 &  & .522\scriptsize$\pm$.043 & \textbf{.680}\scriptsize$\pm$.046 & .668\scriptsize$\pm$.057 & .632\scriptsize$\pm$.048 & .620\scriptsize$\pm$.058 & .676\scriptsize$\pm$.046 & \\
FEVER & .632\scriptsize$\pm$.032 & \cellcolor{stuck}.632\scriptsize$\pm$.032 & \cellcolor{stuck}.632\scriptsize$\pm$.032 & \cellcolor{stuck}.632\scriptsize$\pm$.032 & \cellcolor{stuck}.632\scriptsize$\pm$.032 & \textbf{.726}\scriptsize$\pm$.027 &  & .674\scriptsize$\pm$.038 & .732\scriptsize$\pm$.050 & .736\scriptsize$\pm$.048 & .734\scriptsize$\pm$.044 & \cellcolor{stuck}.674\scriptsize$\pm$.038 & \textbf{.760}\scriptsize$\pm$.028 & \\
DebugBench & .615\scriptsize$\pm$.007 & \cellcolor{stuck}.615\scriptsize$\pm$.007 & \cellcolor{stuck}.615\scriptsize$\pm$.007 & \cellcolor{stuck}.615\scriptsize$\pm$.007 & \cellcolor{stuck}.615\scriptsize$\pm$.007 & \textbf{.666}\scriptsize$\pm$.039 &  & .410\scriptsize$\pm$.020 & .456\scriptsize$\pm$.072 & .510\scriptsize$\pm$.113 & .558\scriptsize$\pm$.128 & \cellcolor{stuck}.410\scriptsize$\pm$.020 & \textbf{.690}\scriptsize$\pm$.049 & \\
\rowcolor{meanrow}\textit{Mean} & .554 & .620 & .628 & .619 & .570 & \textbf{.675} &  & .618 & .667 & .683 & .699 & .634 & \textbf{.747} & \multirow{-7}{*}{\rotatebox[origin=c]{90}{\textbf{Acc.}}} \\
\midrule
GPQA & 1.0\scriptsize$\pm$.0 & \cellcolor{stuck}1.0\scriptsize$\pm$.0 & \cellcolor{stuck}1.0\scriptsize$\pm$.0 & \cellcolor{stuck}1.0\scriptsize$\pm$.0 & \cellcolor{stuck}1.0\scriptsize$\pm$.0 & \textbf{3.2}\scriptsize$\pm$.4 &  & 1.0\scriptsize$\pm$.0 & 1.2\scriptsize$\pm$.4 & 1.2\scriptsize$\pm$.4 & 1.0\scriptsize$\pm$.0 & \cellcolor{stuck}1.0\scriptsize$\pm$.0 & \textbf{1.8}\scriptsize$\pm$.7 & \\
TheoremQA & 1.0\scriptsize$\pm$.0 & 2.0\scriptsize$\pm$.6 & 2.4\scriptsize$\pm$.8 & 1.6\scriptsize$\pm$.5 & 1.2\scriptsize$\pm$.4 & \textbf{3.8}\scriptsize$\pm$.7 &  & 1.0\scriptsize$\pm$.0 & 1.6\scriptsize$\pm$1.2 & 2.2\scriptsize$\pm$1.2 & 2.6\scriptsize$\pm$1.0 & \cellcolor{stuck}1.0\scriptsize$\pm$.0 & \textbf{4.0}\scriptsize$\pm$.6 & \\
HoVer & 1.0\scriptsize$\pm$.0 & \cellcolor{stuck}1.0\scriptsize$\pm$.0 & \cellcolor{stuck}1.0\scriptsize$\pm$.0 & \cellcolor{stuck}1.0\scriptsize$\pm$.0 & \cellcolor{stuck}1.0\scriptsize$\pm$.0 & \textbf{2.4}\scriptsize$\pm$.8 &  & 1.0\scriptsize$\pm$.0 & 1.2\scriptsize$\pm$.4 & 1.2\scriptsize$\pm$.4 & 1.2\scriptsize$\pm$.4 & \cellcolor{stuck}1.0\scriptsize$\pm$.0 & \textbf{2.2}\scriptsize$\pm$1.0 & \\
HotpotQA & 1.0\scriptsize$\pm$.0 & 3.6\scriptsize$\pm$.8 & 3.4\scriptsize$\pm$.8 & 3.8\scriptsize$\pm$.7 & \cellcolor{stuck}1.0\scriptsize$\pm$.0 & \textbf{4.2}\scriptsize$\pm$.7 &  & 1.0\scriptsize$\pm$.0 & \textbf{4.0}\scriptsize$\pm$.6 & 3.6\scriptsize$\pm$.5 & 3.2\scriptsize$\pm$1.0 & 1.8\scriptsize$\pm$.4 & 3.6\scriptsize$\pm$1.0 & \\
FEVER & 1.0\scriptsize$\pm$.0 & \cellcolor{stuck}1.0\scriptsize$\pm$.0 & \cellcolor{stuck}1.0\scriptsize$\pm$.0 & \cellcolor{stuck}1.0\scriptsize$\pm$.0 & \cellcolor{stuck}1.0\scriptsize$\pm$.0 & \textbf{4.4}\scriptsize$\pm$1.0 &  & 1.0\scriptsize$\pm$.0 & 1.8\scriptsize$\pm$.7 & 2.8\scriptsize$\pm$1.0 & 3.0\scriptsize$\pm$1.3 & \cellcolor{stuck}1.0\scriptsize$\pm$.0 & \textbf{3.6}\scriptsize$\pm$.5 & \\
DebugBench & 1.0\scriptsize$\pm$.0 & \cellcolor{stuck}1.0\scriptsize$\pm$.0 & \cellcolor{stuck}1.0\scriptsize$\pm$.0 & \cellcolor{stuck}1.0\scriptsize$\pm$.0 & \cellcolor{stuck}1.0\scriptsize$\pm$.0 & \textbf{3.6}\scriptsize$\pm$.5 &  & 1.0\scriptsize$\pm$.0 & 1.4\scriptsize$\pm$.8 & 1.4\scriptsize$\pm$.5 & 1.8\scriptsize$\pm$.7 & \cellcolor{stuck}1.0\scriptsize$\pm$.0 & \textbf{4.6}\scriptsize$\pm$1.0 & \\
\rowcolor{meanrow}\textit{Mean} & 1.0 & 1.6 & 1.6 & 1.6 & 1.0 & \textbf{3.6} &  & 1.0 & 1.9 & 2.1 & 2.1 & 1.1 & \textbf{3.3} & \multirow{-7}{*}{\rotatebox[origin=c]{90}{\textbf{\#PF}}} \\
\bottomrule
\end{tabular}
}
\end{table*}

\begin{figure*}[t]
\centering
\includegraphics[width=\textwidth]{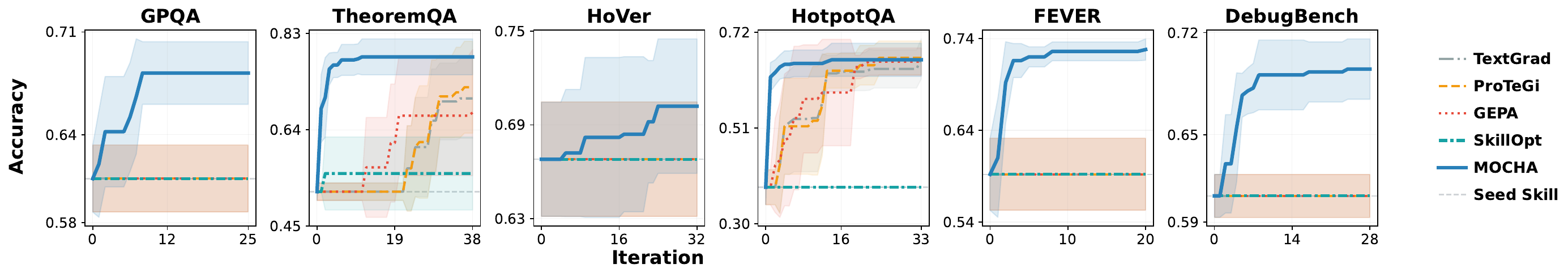}
\caption{\textbf{Optimization dynamics.} Best validation accuracy so far against iteration. On GPQA, HoVer, FEVER, and DebugBench every baseline stays at the seed skill (dashed grey); MOCHA keeps improving.}
\label{fig:evolution}
\end{figure*}

\begin{figure*}[t]
\centering
\includegraphics[width=\textwidth]{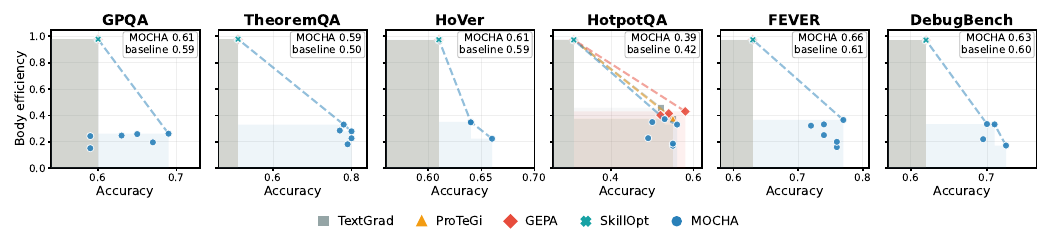}
\caption{\textbf{Accuracy--efficiency fronts of all six skills} (one run each; test split). Shading marks the region each method's front dominates; the box in each panel gives the 2D hypervolume of MOCHA and of the best baseline. The baselines stay at the seed on four skills; MOCHA adds more accurate trade-offs on every skill.}
\label{fig:fronts}
\end{figure*}

\subsection{Main Results}

MOCHA is the only method that improves all three measures at once. Averaged over both models and all six skills, it raises accuracy from $.659$ (GEPA, the strongest baseline) to $.711$ (\textbf{+7.9\%}), hypervolume from $.567$ to $.593$, and the number of Pareto-optimal skills from $1.85$ to $3.45$ (\Cref{tab:main}).

\iflatexml
\begin{figure}[t]
\centering
\includegraphics[width=0.6\linewidth]{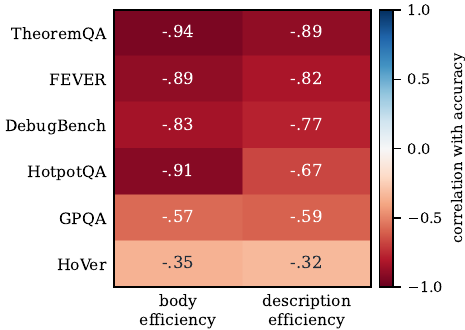}
\caption{\textbf{Accuracy and brevity conflict.} Correlation between accuracy and each efficiency score across all skills found by all methods. More accurate skills are longer on every task.}
\label{fig:conflict}
\end{figure}
\begin{figure}[t]
\centering
\includegraphics[width=0.6\linewidth]{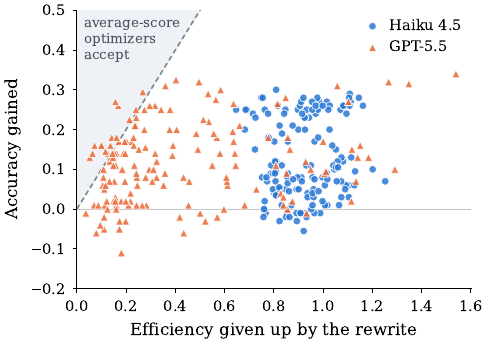}
\caption{\textbf{Why averaging fails.} Each point is a skill MOCHA kept, compared with its seed: accuracy gained against total efficiency given up. An optimizer that maximizes the average score accepts only skills in the shaded region. Most of MOCHA's skills are more accurate than the seed, yet almost none would pass.}
\label{fig:breakeven}
\end{figure}
\else
\begin{figure*}[t]
\centering
\begin{minipage}[b]{0.40\textwidth}
    \centering
    \includegraphics[width=\linewidth]{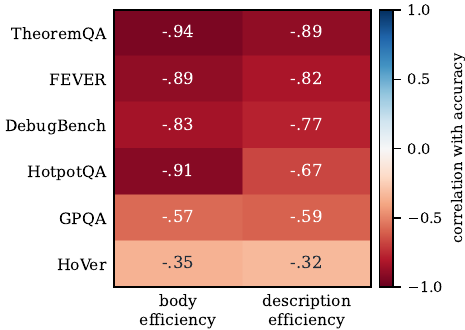}
    \captionof{figure}{\textbf{Accuracy and brevity conflict.} Correlation between accuracy and each efficiency score across all skills found by all methods. More accurate skills are longer on every task.}
    \label{fig:conflict}
\end{minipage}\hfill
\begin{minipage}[b]{0.54\textwidth}
    \centering
    \includegraphics[width=0.75\linewidth]{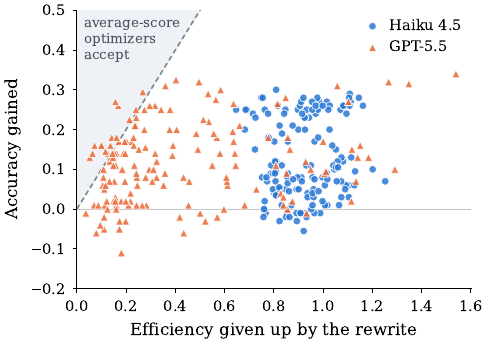}
    \captionof{figure}{\textbf{Why averaging fails.} Each point is a skill MOCHA kept, compared with its seed: accuracy gained against total efficiency given up. An optimizer that maximizes the average score accepts only skills in the shaded region. Most of MOCHA's skills are more accurate than the seed, yet almost none would pass.}
    \label{fig:breakeven}
\end{minipage}
\end{figure*}
\fi
\textbf{Baselines get stuck.} With Claude Haiku~4.5, every baseline returns the seed skill unchanged in every run on GPQA, HoVer, FEVER, and DebugBench (red cells in \Cref{tab:main}); their curves in \Cref{fig:evolution} stay flat at the seed for the whole budget. Across both models, TextGrad, ProTeGi, and GEPA return the seed in 112 of their 180 runs and SkillOpt in 50 of 55, whereas MOCHA moves beyond it in 58 of its 60 runs. The stall is not a defect of the baselines' setup: they receive the same rewrite prompt and per-objective feedback as MOCHA, the rewrites they are offered are just as long and as accurate, and they \emph{do} improve whenever the seed is weak (HotpotQA, TheoremQA). \Cref{sec:why} shows that their acceptance rule, not their proposals, holds them back.

\textbf{MOCHA breaks through.} MOCHA has the highest mean accuracy with both models: $.675$ vs.\ $.628$ for ProTeGi with Claude Haiku~4.5 (\textbf{+7.5\%}, and +21.8\% over the seed) and $.747$ vs.\ $.699$ for GEPA with GPT 5.5 (\textbf{+6.9\%}). The largest gains are on FEVER with Claude Haiku~4.5 (\textbf{+14.9\%}), where every baseline stalls, and on DebugBench with GPT 5.5 (\textbf{+23.7\%}, $.690$ vs.\ $.558$), and MOCHA also leads where the baselines do move (TheoremQA: +10.4\% with Claude Haiku~4.5). It trails in only three of the twelve skill--model pairs, each within one standard deviation and each as the analysis in \Cref{sec:why} predicts: on HotpotQA the seed is weak ($.336$ and $.522$), so simple formatting rules lift every optimizer, and with GPT 5.5 the GPQA seed is already near the model's ceiling ($.860$).

\textbf{Better fronts.} MOCHA has the highest mean hypervolume with both models ($.531$ vs.\ $.515$ and $.655$ vs.\ $.625$) and finds up to \textbf{twice} as many Pareto-optimal skills (3.6 vs.\ at most 1.6 with Haiku; 3.3 vs.\ 2.1 with GPT 5.5). Baselines that stall keep a single point, the seed; MOCHA keeps the seed and adds more accurate trade-offs (\Cref{fig:fronts}). The one exception is HotpotQA, where the baselines also move and their fronts are competitive. The absolute HV gains look modest because the seed, the most efficient skill, lies on every front; HV differences therefore measure what each method adds on the accuracy side of the trade-off.

\textbf{The gains are statistically reliable.} Every method sees the same data splits for a given seed, so we compare MOCHA with each baseline over the 30 paired (skill, seed) runs with a Wilcoxon signed-rank test. With Claude Haiku~4.5, MOCHA is more accurate than ProTeGi in 21 of 30 runs and than each baseline at $p \le .002$; with GPT 5.5, it beats each baseline at $p \le .03$. MOCHA also varies less across seeds where every method moves (TheoremQA: $\pm.04$ vs.\ $\pm.06$--$.13$).

\textbf{The deployed skills stay within the limits.} MOCHA's deployed skills use more of their length budget than the baselines', which mostly return the seed: their description and body efficiency average $.75$ and $.47$, against $.85$ and $.82$ (\Cref{tab:compliance_app} in \Cref{app:compliance}). All remain within both limits, and the hypervolume, which counts all three objectives, still favors MOCHA.

\textbf{All methods spend the same budget.} MOCHA keeps 4.9 skills per run with Claude Haiku~4.5 and 5.3 with GPT 5.5, against at most 0.8 and 1.7 for the baselines. Each kept skill costs a validation pass, so MOCHA spends more of the same 1000-rollout budget on validation and less on proposals; its gains do not come from extra compute.

\subsection{Why Do Baselines Get Stuck?}
\label{sec:why}

The answer lies in the tension between the objectives. Across every task, more accurate skills are longer (\Cref{fig:conflict}): the extra accuracy comes from procedures, checks, and output rules that the one-line seeds lack. An optimizer that judges a rewrite by its average score therefore accepts it only when the accuracy it gains exceeds the efficiency it gives up. For a short seed, that bar is high. The conflict is not occasional: of the 453 kept skills, across all methods and both models, that are more accurate than their seed, every single one is less efficient.

\begin{table}[t]
\centering
\caption{\textbf{Exploration versus exploitation.} MOCHA's row gives its HV and accuracy; the other rows give the difference from MOCHA.}
\label{tab:ablation}
\footnotesize
\setlength{\tabcolsep}{6pt}
\begin{tabular}{@{}lcc@{}}
\toprule
\textbf{Variant} & \textbf{HV}$\uparrow$ & \textbf{Acc.}$\uparrow$ \\
\midrule
w/o HVC \scriptsize{(exploit only)} & \cellcolor{ablminus}$-.001$ & \cellcolor{ablplus}$+.012$ \\
w/o annealing \scriptsize{(explore only)} & \cellcolor{ablplus}$+.002$ & \cellcolor{ablminus}$-.004$ \\
\textbf{MOCHA} \scriptsize{(explore, then exploit)} & \cellcolor{ablbase}\textbf{.531} & \cellcolor{ablbase}\textbf{.675} \\
\bottomrule
\end{tabular}
\end{table}

\begin{figure*}[!t]
\centering
\begin{skillbox}[colback=gray!4, colframe=gray!50!black,
  fonttitle=\small, boxrule=0.4pt, left=3pt, right=3pt, top=2pt, bottom=2pt,
  arc=2pt]{\small\bfseries FEVER Claim Verification --- SKILL.md}
\scriptsize\ttfamily
\textcolor{gray}{---}\\
\textcolor{blue!70}{name:} fever\_verification \quad
\textcolor{blue!70}{compatibility:} dspy>=2.5, python>=3.10\\
\textcolor{blue!70}{metadata:} task: fever\_singlecall, modules: 1 \quad
\textcolor{blue!70}{allowed-tools:} [retriever]\\
\textcolor{gray}{---}\\[3pt]
\colorbox{red!8}{\parbox{\dimexpr\linewidth-6pt\relax}{
\scriptsize\ttfamily
\textcolor{blue!70}{description:} Claim verification with retrieval tool
\hfill\textnormal{desc.\ eff.\,=\,0.93}\\[1pt]
\textcolor{gray}{\# execute.predict}\\
Given the fields `claim', `evidence', produce the fields `verdict'.
\hfill\textnormal{body eff.\,=\,0.97}
}}\\[1pt]
\hfill\textcolor{red!60!black}{\textnormal{\scriptsize\bfseries Seed (returned by all baselines): accuracy 0.63 | desc.\ eff.\ 0.93 | body eff.\ 0.97 | average 0.84}}\\[4pt]
\colorbox{green!8}{\parbox{\dimexpr\linewidth-6pt\relax}{
\scriptsize\ttfamily
\textcolor{blue!70}{description:} FEVER-style 3-class claim verification: classify as SUPPORTS,\\
\hspace*{2em}REFUTES, or NOT ENOUGH INFO based strictly on provided evidence passages.
\hfill\textnormal{desc.\ eff.\,=\,0.77}\\[1pt]
\textcolor{gray}{\# execute.predict}\\
Given `claim' and `evidence', produce `verdict'. Must be: SUPPORTS, REFUTES, or\\NOT ENOUGH INFO.\\[1pt]
\textbf{Rule 1}: Strict evidence-only reasoning. No outside knowledge.\\
\textbf{Rule 2}: ALL claim parts must be supported for SUPPORTS.\\
\textbf{Rule 3}: REFUTES requires direct contradiction only.\\
\textbf{Rule 4}: NOT ENOUGH INFO is default when evidence is partial or about related entities.\\
\textbf{Rule 5}: Never invent labels.\\[1pt]
\textbf{Step-by-step}: (1) Decompose claim $\to$ (2) Check each assertion $\to$ (3) Check contradiction $\to$ (4) Aggregate
\hfill\textnormal{body eff.\,=\,0.38}
}}\\[1pt]
\hfill\textcolor{green!50!black}{\textnormal{\scriptsize\bfseries MOCHA: accuracy 0.70 | desc.\ eff.\ 0.77 | body eff.\ 0.38 | average 0.62.
 (Body shortened for display; the full skill has 51 lines.)}}
\end{skillbox}
\caption{\textbf{FEVER skill before and after optimization.} Red: the seed skill, which every baseline returns unchanged. Green: a skill found by MOCHA (body shortened for display). The new skill is more accurate but longer, so its average score falls: an optimizer that maximizes the average rejects it, whereas MOCHA keeps it as a new trade-off.}
\label{fig:skill_compare}
\end{figure*}

\begin{figure}[t]
    \centering
    \includegraphics[width=0.8\linewidth]{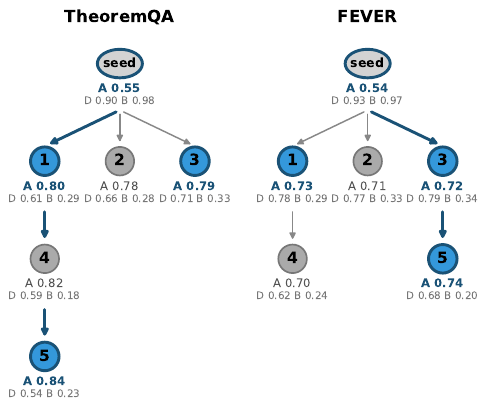}
    \caption{\textbf{Skill evolution trees} of two MOCHA runs. Nodes are kept skills in order of discovery, labelled with accuracy (A) and description/body efficiency (D/B); blue marks the skills on the Pareto front and grey the dominated ones.}
    \label{fig:tree}
    \vspace{-0.25cm}
\end{figure}
\Cref{fig:breakeven} makes this concrete. It plots every skill MOCHA kept against the seed it started from. Most gain accuracy, but nearly all give up more efficiency than they gain: none of the 147 skills found with Claude Haiku~4.5 would pass an average-score test, and only 15 of 159 with GPT 5.5, whose rewrites are shorter. This is why the baselines stall: their rule is working as designed, but it cannot see that a longer, more accurate skill is a useful trade-off.

\subsection{Ablation Study}
\label{sec:ablation}

\Cref{tab:ablation} compares MOCHA with its two extremes: exploiting from the start (w/o HVC) and exploring throughout (w/o annealing). They pull in opposite directions. Exploiting alone finds slightly more accurate skills but a smaller front; exploring alone finds a larger front but less accurate skills. MOCHA, which explores first and exploits later, lies between them on both, which is the balance it is designed to strike and why it is our default. \Cref{fig:tree} shows this ordering at work in two runs: during exploration, MOCHA branches out from the seed into several distinct rewrites, each a large jump in accuracy; during exploitation, it refines one of these branches further.

\section{Discussion and Conclusion}
\label{sec:discussion}

\textbf{Pareto front, not a single skill.} Both runs in \Cref{fig:tree} show how exploration and exploitation build a set of useful trade-offs. On TheoremQA, exploration raises accuracy from 55\% to about 80\% along several branches, and exploitation extends one branch to 84\%. The final front keeps the seed, the most accurate skill, and node~3, which reaches 79\% with a 13\% shorter body. FEVER shows the same pattern: the final front includes the seed and three rewrites (nodes 1, 3, and 5), each balancing accuracy and efficiency differently. These skills are all non-dominated, so the best deployment choice depends on which trade-off matters; no optimizer can choose it in advance. Evolution trees for all six skills are in the appendix (\Cref{app:evolution_trees}).

\textbf{What the extra length buys.} MOCHA's skills are longer than the baselines' seeds (\Cref{tab:compliance_app} in \Cref{app:compliance}). The extra length lets them give complete instructions: how to break down a claim, check every answer option, or format an answer—steps the one-line seeds leave unstated (\Cref{fig:skill_compare}; more examples in \Cref{app:qualitative}). The advantage is not simply that MOCHA proposes longer rewrites. All methods use the same rewrite prompt and are offered the same long, more accurate rewrites; the baselines reject them, while MOCHA keeps them.

\textbf{Conclusion.} Skill optimization is inherently multi-objective: platform limits on description and body length create trade-offs that single-objective optimizers cannot see. Our central finding is that existing optimizers, which fold these objectives into one score, make no progress in most of their runs: 1000 rollouts yield zero improvement over the seed. The cause is structural: every useful rewrite gains accuracy at the cost of efficiency and averaging rejects it. MOCHA breaks through this barrier by selecting with Chebyshev scalarization and accepting whatever expands the Pareto front, achieving 7.9\% higher mean accuracy than the strongest baseline, with gains up to 23.7\% on DebugBench with GPT 5.5 and 14.9\% on FEVER with Claude Haiku~4.5, while returning nearly twice as many Pareto-optimal skills for practitioners to choose from.

\FloatBarrier
\section*{Limitations}

\textbf{Objectives.} We measure efficiency by length against platform limits. Deployments may also care about latency, cost, or tool schemas; MOCHA applies to them directly once each is expressed as a metric in $[0,1]$. This normalization keeps Chebyshev scores comparable across objectives and requires a ceiling for each metric, which is natural for bounded quantities such as limits and budgets and only needs to be chosen once for unbounded ones.

\textbf{Skill versus harness optimization.} We optimize the text of skills while keeping the execution harness fixed. Searching the harness itself, such as its control flow and tool use, jointly with its skills~\citep{lee2026meta} is a natural extension, since MOCHA needs only metrics and a rewrite operator.

\textbf{Choosing from the front.} For comparison with single-objective baselines, we report the most accurate skill on validation and summarize the rest of the front through hypervolume and front size. Helping practitioners choose among the returned skills, for example through preference elicitation, is left to future work.

\textbf{Weak seeds.} When the seed is weak, simpler optimizers also find large gains, and MOCHA's added value lies mainly in the front it returns (HotpotQA). Predicting in advance how much a seed stands to gain is an interesting direction.

\section*{Ethical Considerations}

MOCHA refines the text of existing skills through API access only and introduces no new model capabilities. Optimized skills could in principle serve harmful tasks, a risk shared by all prompt optimization methods; MOCHA's outputs are human-readable and can be audited before deployment. All datasets are public and used under their licenses (\Cref{app:reproducibility}). To support reproducibility, the appendix includes Python-like pseudocode of the full algorithm (\Cref{app:pseudocode}) and an \texttt{AGENTS.md} guide for re-implementing MOCHA and reproducing our experiments (\Cref{app:agents}).

\bibliography{references}

\clearpage
\appendix

\section{Background: Scalarization and Hypervolume Theory}
\label{app:background}

We provide background on the multi-objective tools underlying MOCHA.

\subsection{Multi-Objective Optimization}

In general no single skill is best on every objective at once. Two paradigms exist~\citep{emmerich2018tutorial}: \emph{a-priori} methods, where the decision maker's preferences are known in advance, and \emph{a-posteriori} methods, which learn the Pareto front for later selection. MOCHA is an a-posteriori method.

\subsection{Linear vs.\ Chebyshev Scalarization}

\emph{Scalarization} reduces a multi-objective problem to a single objective through a weight vector $\vw$:
\begin{align}
  \text{Linear:} \quad s_\vw^{\mathrm{lin}}(p) &= \textstyle\sum_{i=1}^M w_i \, m_i(p), \label{eq:linear_app} \\
  \text{Chebyshev:} \quad s_\vw(p) &= 1 - \max_{j} \, w_j \, \big(1 - m_j(p)\big). \label{eq:cheb_app}
\end{align}
Linear scalarization picks the skill with the largest weighted sum; Chebyshev picks the skill with the largest score, i.e., the smallest worst-case weighted gap to the ideal point $(1, \ldots, 1)$. Linear scalarization misses Pareto-optimal points in non-convex regions of the front, while Chebyshev can reach \emph{every} Pareto-optimal point~\citep{miettinen1999nonlinear}. We sample $\vw$ uniformly from the simplex ($\vw \sim \mathrm{Dirichlet}(\mathbf{1})$), which treats all objectives symmetrically and needs no tuning.

\subsection{Hypervolume Indicator}
\label{app:hv}

The hypervolume indicator~\citep{guerreiro2021hypervolume}, also called the $S$-metric~\citep{zitzler1999multiobjective}, is the Lebesgue measure $\lambda$ of the region dominated by a set $\cP$ and bounded by a reference point $r$:
\begin{equation}
  \hv(\cP) = \lambda\Big(\textstyle\bigcup_{p \in \cP} \prod_{i=1}^M [r_i,\, m_i(p)]\Big).
  \label{eq:hv_app}
\end{equation}
All three objectives (accuracy, description efficiency, body efficiency) are non-negative and maximized, so we use $r = \mathbf{0}$. The hypervolume contribution of a skill, $\hvc(p, \cP) = \hv(\cP \cup \{p\}) - \hv(\cP)$ (\Cref{eq:hvc}), is positive if and only if no member of $\cP$ dominates $p$.

\textbf{Key properties.}
\begin{itemize}[nosep,leftmargin=*]
  \item \emph{Strict monotonicity}: if $\cP'$ Pareto-dominates $\cP$, then $\hv(\cP') > \hv(\cP)$; hypervolume is the only known unary indicator with this property~\citep{zitzler2003performance}.
  \item \emph{Computation}: exact HV is NP-hard in the number of objectives~\citep{bringmann2010approximation}, but fast for three~\citep{guerreiro2021hypervolume}.
  \item \emph{Complementarity with Chebyshev scalarization}: Chebyshev scalarization targets one direction of the front for a given $\vw$, while HVC rewards a candidate that expands the front in \emph{any} direction. MOCHA uses the second to explore and the first to exploit.
\end{itemize}

\subsection{GEPA Framework}

GEPA~\citep{agrawal2025gepa} optimizes text through iterative evolution: it evaluates candidates, gathers textual feedback, selects promising candidates, and mutates them. MOCHA keeps this mutation and evaluation loop and replaces candidate selection and acceptance with the multi-objective mechanisms of \Cref{sec:method}.

\section{Implementation Details}
\label{app:implementation}

\textbf{Two-stage evaluation.} (1)~\emph{Minibatch gating}: the parent and the candidate are evaluated on a minibatch of $n{=}10$ training examples, and the acceptance rule (HVC or Chebyshev) is applied to the minibatch scores, which filters poor candidates cheaply. (2)~\emph{Validation scoring}: accepted candidates are evaluated on the 100 validation examples and committed to the pool; these scores drive parent selection.

\textbf{Budget accounting.} The budget $B$ counts rollouts, i.e., metric calls: one per example scored, on a minibatch or on the validation set~\citep{agrawal2025gepa}. An iteration costs $2n$ rollouts, plus $|\cD_{\mathrm{val}}|$ if a candidate is committed.

\textbf{Scheduler and annealing.} During exploration, a candidate is kept when its hypervolume contribution over its parent exceeds a threshold $\alpha(b) = \alpha_0 e^{-\lambda b/B}$, where $b$ is the budget spent. We use $\alpha_0 = 0.1$ and $\lambda = 10$; once $\alpha(b)$ falls below $10^{-3}$, after about 46\% of the budget, MOCHA switches to the Chebyshev rule (\Cref{fig:schedule}). Removing HVC corresponds to exploiting from the start; removing annealing keeps the threshold above the switching level, so MOCHA explores throughout.

\textbf{Buffer.} During exploration, candidates that expand the front but fall below the threshold wait in a buffer of the $K{=}5$ best by hypervolume contribution. The best buffered candidate is committed once it clears the threshold.

\textbf{Matched evaluation setup.} TextGrad (greedy acceptance), ProTeGi (UCB beam search, width 3), GEPA (Pareto selection over validation examples), and MOCHA share the same rewrite prompt, evaluation harness, and rollout budget, and differ only in how they select and accept candidates; the baselines accept a candidate when its average objective on the minibatch improves. SkillOpt~\citep{yang2026skillopt} accepts an edit when the average objective improves on the validation set, and uses the same seed skills and data splits. For every method, the deployed skill is the first committed skill with the highest validation accuracy.

\subsection{Detailed Algorithm}
\label{app:pseudocode}

\Cref{lst:mocha} gives MOCHA in Python-like pseudocode, including the budget accounting, the exploration threshold, and the buffer omitted from \Cref{alg:main}. Metrics are vectors in $[0,1]^M$, all maximized.

\begin{figure*}[t]
\begin{lstlisting}[style=pseudo]
from <any multi-objective library> import hypervolume
#   any exact HV routine works; in Python, e.g. the HV indicator of pymoo
#   (pymoo minimizes, so pass negated scores)

def cheb(m, w):                              # Chebyshev score, higher is better
    return 1 - max(w[j] * (1 - m[j]) for j in range(M))

def hv(points):                              # all objectives maximized
    return hypervolume(points, reference=zeros(M))

def mocha(seed, train, val, budget=1000, batch_size=10,
          alpha0=0.1, decay=10.0, eps=1e-3, K=5):
    pool = [(seed, evaluate(seed, val))]      # (skill, validation scores)
    spent = len(val)                         # the seed's validation pass
    buffer = []
    while spent + 2 * batch_size <= budget:
        w = dirichlet(ones(M))               # parent selection: random weighting
        parent, parent_val = max(pool, key=lambda e: cheb(e[1], w))
        batch = sample(train, batch_size)
        before = mean_scores(parent, batch)
        child = rewrite(parent, feedback(parent, batch))
        after = mean_scores(child, batch)
        spent += 2 * batch_size
        progress = (spent - len(val)) / (budget - len(val))
        alpha = alpha0 * exp(-decay * progress)
        chosen = None
        if alpha > eps:                      # pool expansion: explore
            gain = hv([before, after]) - hv([before])   # HVC over the parent
            if gain > 0:
                buffer = top_k(buffer + [(gain, child)], K)
            if buffer and buffer[0][0] > alpha:
                chosen = buffer.pop(0)[1]
        elif cheb(after, w) > cheb(before, w):   # pool expansion: exploit
            chosen = child
        if chosen is not None and spent + len(val) <= budget:
            pool.append((chosen, evaluate(chosen, val)))
            spent += len(val)
    return pareto_front(pool)

\end{lstlisting}
\caption{\textbf{MOCHA in Python-like pseudocode.}}
\label{lst:mocha}
\end{figure*}

Here \texttt{evaluate} and \texttt{mean\_scores} return the mean of each metric over a set of examples, \texttt{rewrite} calls the LLM with the prompt of \Cref{app:mutation}, \texttt{hv} computes the exact hypervolume with reference point $\mathbf{0}$ (any standard implementation can be used), and \texttt{top\_k} keeps the $K$ entries with the largest gain. The single-objective baselines replace both acceptance rules by ``keep the child if \texttt{mean(after) > mean(before)}'' and use their own rule for parent selection.

\subsection{Reproduction Guide}
\label{app:agents}

The guide in \Cref{lst:agents}, written as an \texttt{AGENTS.md} file, is meant to let a person or a coding agent re-implement MOCHA and reproduce our experiments from this paper alone.

\begin{figure*}[t]
\begin{lstlisting}[style=guide]
# AGENTS.md - Reproducing MOCHA

## Goal
Implement MOCHA (Algorithm 1, Appendix B.1) and the baselines, run them on six skills with two LLMs, and reproduce Tables 1-3.

## 1. Components to implement
1. Skill: YAML front matter (name, description, metadata) + Markdown body. Only `description` and `body` are optimized.
2. Executor: runs a task example with the skill text in the system prompt and returns the model's answer. One scored example (a metric call) = one rollout, on minibatch and validation examples alike.
3. Metrics, all in [0, 1], higher is better:
   - accuracy: exact match (GPQA, HoVer, FEVER, TheoremQA after numeric normalization), token F1 (HotpotQA), test pass rate (DebugBench);
   - efficiency of a field = max(0, 1 - len(field) / limit), with limit = 1024 chars (description) and 5000 chars (body).
4. Rewrite operator: one LLM call with the prompt in Appendix C.3, given the current skill, per-example feedback on a minibatch, and each field's length vs. its limit; parse the returned SKILL.md.
5. Hypervolume: exact 3-D hypervolume, reference point (0, 0, 0). Any standard implementation works (e.g., in Python, pymoo's HV indicator applied to negated scores, since pymoo assumes minimization).
6. MOCHA loop: Appendix B.1. Baselines: same loop and rewrite operator, accepting a child only if its mean objective beats the parent's.

## 2. Configuration (identical for all tasks and models)
budget = 1000 rollouts; batch_size = 10; splits 100/100/100 (train/val/test), resampled per seed; seeds = 0, 42, 123, 456, 786; alpha0 = 0.1; decay = 10; eps = 1e-3; buffer K = 5.

## 3. Running
For each model, skill, method, and seed:
  1. Sample the splits with the seed.
  2. Run the optimizer until the budget is spent.
  3. Log every kept skill with its validation and test scores.
Test scores are logged for analysis only; never use them for selection.

## 4. Reporting (Table 1)
- Accuracy: deploy the FIRST kept skill with the highest validation accuracy; report its test accuracy. Mean +- std over the five seeds.
- HV: hypervolume of the test scores of all kept skills of a run.
- #PF: number of those skills that are non-dominated on test scores.
- Average each metric over seeds per skill, then over skills.

## 5. Sanity checks
- With averaged scores, the baselines should rarely move beyond a one-line seed on GPQA, HoVer, FEVER, and DebugBench.
- No reported skill should exceed a field limit.
- MOCHA should keep several skills per run (about 5 in our runs).
\end{lstlisting}
\caption{\textbf{\texttt{AGENTS.md}: implementing and reproducing MOCHA.}}
\label{lst:agents}
\end{figure*}

\section{Additional Experimental Results}
\label{app:additional}

\subsection{Hyperparameter Sensitivity}
\label{app:sensitivity}

We vary the buffer size $K$, the initial threshold $\alpha_0$, and the decay rate $\lambda$ one at a time (\Cref{tab:sensitivity}). Averaged over datasets, accuracy stays within $.72$--$.77$ across all settings, so the reported results do not depend on a finely tuned configuration.

\begin{table}[h]
\centering
\footnotesize
\setlength{\tabcolsep}{3.5pt}
\caption{\textbf{Sensitivity}: MOCHA test accuracy under one-at-a-time sweeps, averaged over datasets. Default: $K{=}5$, $\alpha_0{=}0.1$, $\lambda{=}10$.}
\label{tab:sensitivity}
\resizebox{\linewidth}{!}{\begin{tabular}{@{}lccccccc@{}}
\toprule
 & default & $K{=}1$ & $\alpha_0{=}.05$ & $\alpha_0{=}.2$ & $\alpha_0{\to}0$ & $\lambda{=}5$ & $\lambda{=}15$ \\
\midrule
Accuracy & .745 & .740 & .745 & .720 & .765 & .765 & .745 \\
\bottomrule
\end{tabular}}
\end{table}

\subsection{Buffer Size}
\label{app:buffer}

\Cref{tab:buffer} compares a buffer of one candidate with the default of five. Holding the best five exploratory candidates lets MOCHA spend each validation pass on a stronger skill, which raises both hypervolume and accuracy; we therefore use $K{=}5$.

\begin{table}[h]
\centering
\caption{\textbf{Buffer size}, averaged over datasets.}
\label{tab:buffer}
\footnotesize
\setlength{\tabcolsep}{5pt}
\begin{tabular}{@{}lccc@{}}
\toprule
\textbf{Variant} & \textbf{HV} & \textbf{Acc.} & \textbf{\#PF} \\
\midrule
$K{=}1$ & .636 & .725 & \textbf{3.4} \\
$K{=}5$ (default) & \textbf{.655} & \textbf{.747} & 3.3 \\
\bottomrule
\end{tabular}
\end{table}

\subsection{Shared Mutation Interface}
\label{app:mutation}

TextGrad, ProTeGi, GEPA, and MOCHA share the rewrite prompt below. \texttt{\{compliance\_report\}} lists each field's length against its limit (e.g., \texttt{body: 6,412/5,000 chars}); \texttt{\{feedback\_text\}} gives per-example feedback such as \texttt{Incorrect. Expected 'SUPPORTS', got 'REFUTES'.}

\begin{skillbox}[colback=blue!2, colframe=blue!40!black,
  fonttitle=\small, boxrule=0.5pt, left=5pt, right=5pt, top=4pt, bottom=4pt,
  arc=3pt, fontupper=\footnotesize\ttfamily]{\small\bfseries Rewrite prompt (shared by TextGrad, ProTeGi, GEPA, and MOCHA)}
You are optimizing a SKILL.md specification for a language model skill.\\[6pt]
\textcolor{blue!60!black}{\bfseries\#\# SKILL.md Format Constraints}\\
- description: \(\le\)1,024 characters\\
- body: \(\le\)5,000 characters\\
- All YAML frontmatter fields must remain valid YAML\\[6pt]
\textcolor{blue!60!black}{\bfseries\#\# Current SKILL.md}\\
\{current\_skill\_md\}\\[6pt]
\textcolor{blue!60!black}{\bfseries\#\# Compliance Status}\\
\{compliance\_report\}\\[6pt]
\textcolor{blue!60!black}{\bfseries\#\# Task Examples with Feedback}\\
\{feedback\_text\}\\[6pt]
\textcolor{blue!60!black}{\bfseries\#\# Sections to Improve:} \{components\_to\_update\}\\[6pt]
Rewrite the SKILL.md: TOP priority is to improve task accuracy by adding as many steps or instructions as necessary while still respecting all field constraints. You may modify ANY field (description, body sections). Maintain valid YAML frontmatter and \textasciigrave\# section\_name\textasciigrave\ headers in the body.\\[4pt]
Return the complete SKILL.md within \textasciigrave\textasciigrave\textasciigrave\ blocks.
\end{skillbox}

The prompt asks for accuracy first and reports lengths, so every method receives the same pressure toward longer, more accurate skills; only selection and acceptance differ.

\subsection{Per-Task Efficiency Analysis}
\label{app:compliance}

\Cref{tab:compliance_app} reports the description and body efficiency of the deployed skills for all methods across the six skills.

\begin{table}[h]
\centering
\caption{\textbf{Token efficiency} of the deployed skills (mean over five seeds). Desc.\ = description ($\le$1,024 characters); Body = instruction body ($\le$5,000 characters).}
\label{tab:compliance_app}
\footnotesize
\setlength{\tabcolsep}{3pt}
\resizebox{\linewidth}{!}{\begin{tabular}{@{}l cc cc cc cc cc@{}}
\toprule
& \multicolumn{2}{c}{\textbf{TextGrad}} & \multicolumn{2}{c}{\textbf{ProTeGi}} & \multicolumn{2}{c}{\textbf{GEPA}} & \multicolumn{2}{c}{\textbf{SkillOpt}} & \multicolumn{2}{c}{\textbf{MOCHA}} \\
\cmidrule(lr){2-3} \cmidrule(lr){4-5} \cmidrule(lr){6-7} \cmidrule(lr){8-9} \cmidrule(lr){10-11}
\textbf{Skill} & Desc. & Body & Desc. & Body & Desc. & Body & Desc. & Body & Desc. & Body \\
\midrule
GPQA & .95 & .99 & .95 & .99 & .95 & .99 & .95 & .99 & .72 & .25 \\
TheoremQA & .73 & .46 & .72 & .47 & .77 & .59 & .95 & .96 & .68 & .32 \\
HoVer & .96 & .99 & .96 & .99 & .96 & .99 & .96 & .99 & .77 & .34 \\
HotpotQA & .82 & .42 & .81 & .39 & .79 & .44 & .96 & .99 & .82 & .38 \\
FEVER & .96 & .99 & .96 & .99 & .96 & .99 & .96 & .99 & .78 & .31 \\
DebugBench & .95 & .98 & .95 & .98 & .95 & .98 & .95 & .98 & .69 & .36 \\
\midrule
\textbf{Mean} & .90 & .81 & .89 & .80 & .90 & .83 & .96 & .98 & .74 & .33 \\
\bottomrule
\end{tabular}}
\end{table}

The baselines keep high efficiency because they rarely accept a candidate that departs from the short seed. MOCHA trades efficiency for accuracy, and its multi-objective selection makes this trade-off explicit rather than hidden.

\subsection{Full Pareto Front Visualization}
\label{app:pareto_2d}

\begin{figure*}[h]
\centering
\includegraphics[width=\textwidth]{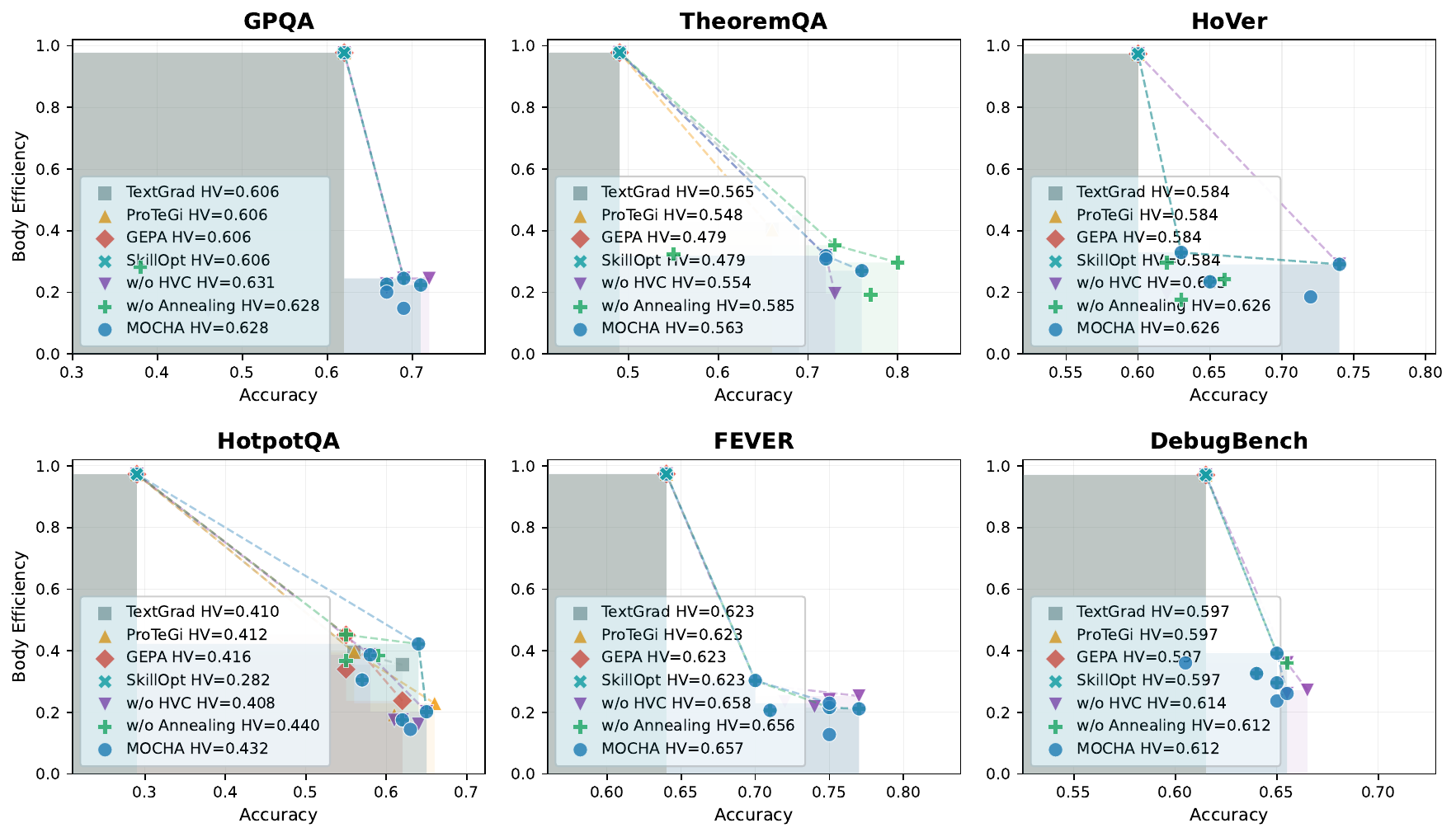}
\caption{\textbf{2D Pareto fronts} (accuracy $\times$ body efficiency) for all six skills, with the baselines and three MOCHA variants. Shaded regions mark dominated hypervolume.}
\label{fig:pareto_2d_body}
\end{figure*}

\begin{figure*}[h]
\centering
\includegraphics[width=\textwidth]{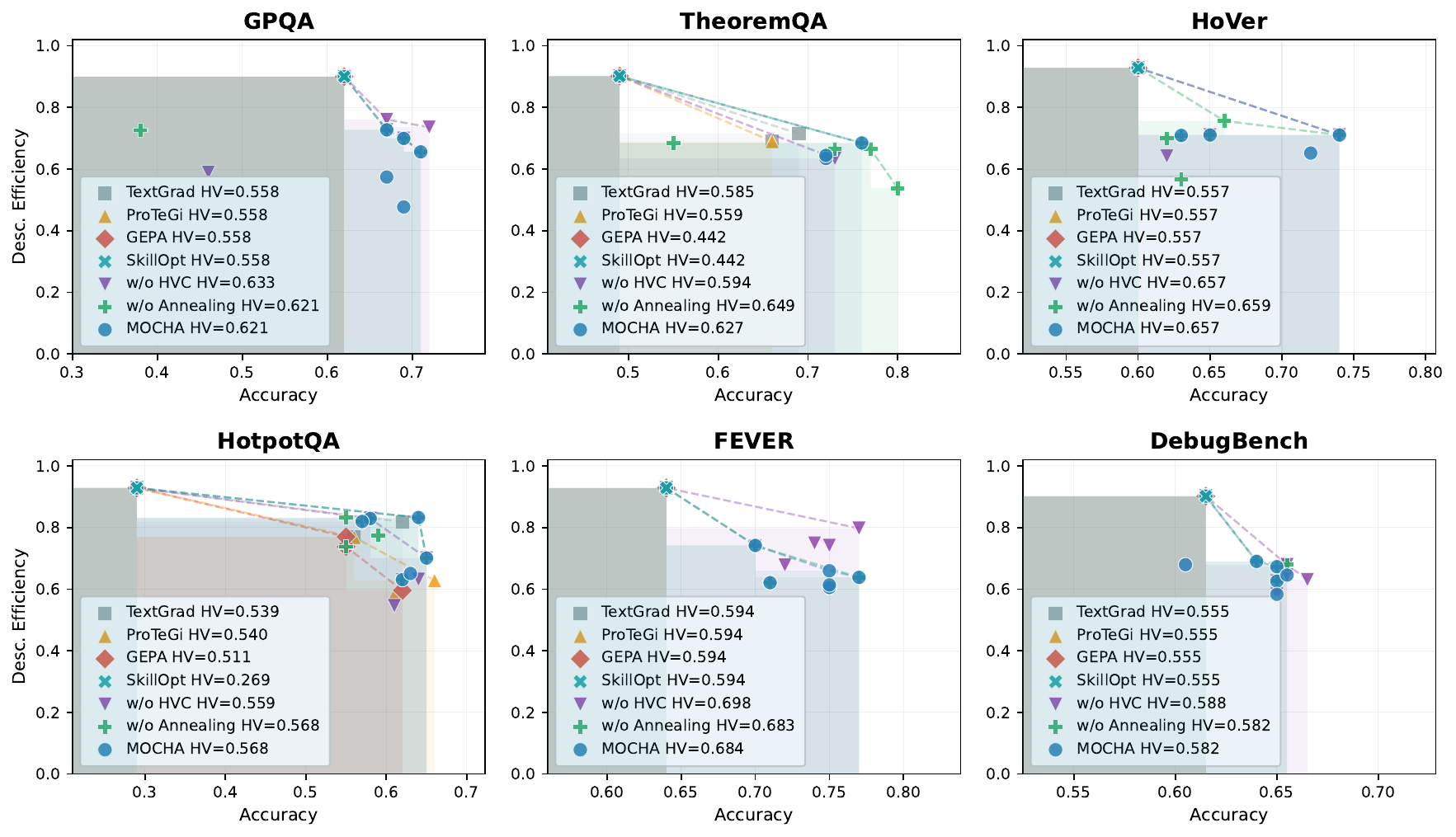}
\caption{\textbf{2D Pareto fronts} (accuracy $\times$ description efficiency) for all six skills. The same pattern holds.}
\label{fig:pareto_2d_desc}
\end{figure*}

\begin{figure*}[h]
\centering
\includegraphics[width=\textwidth]{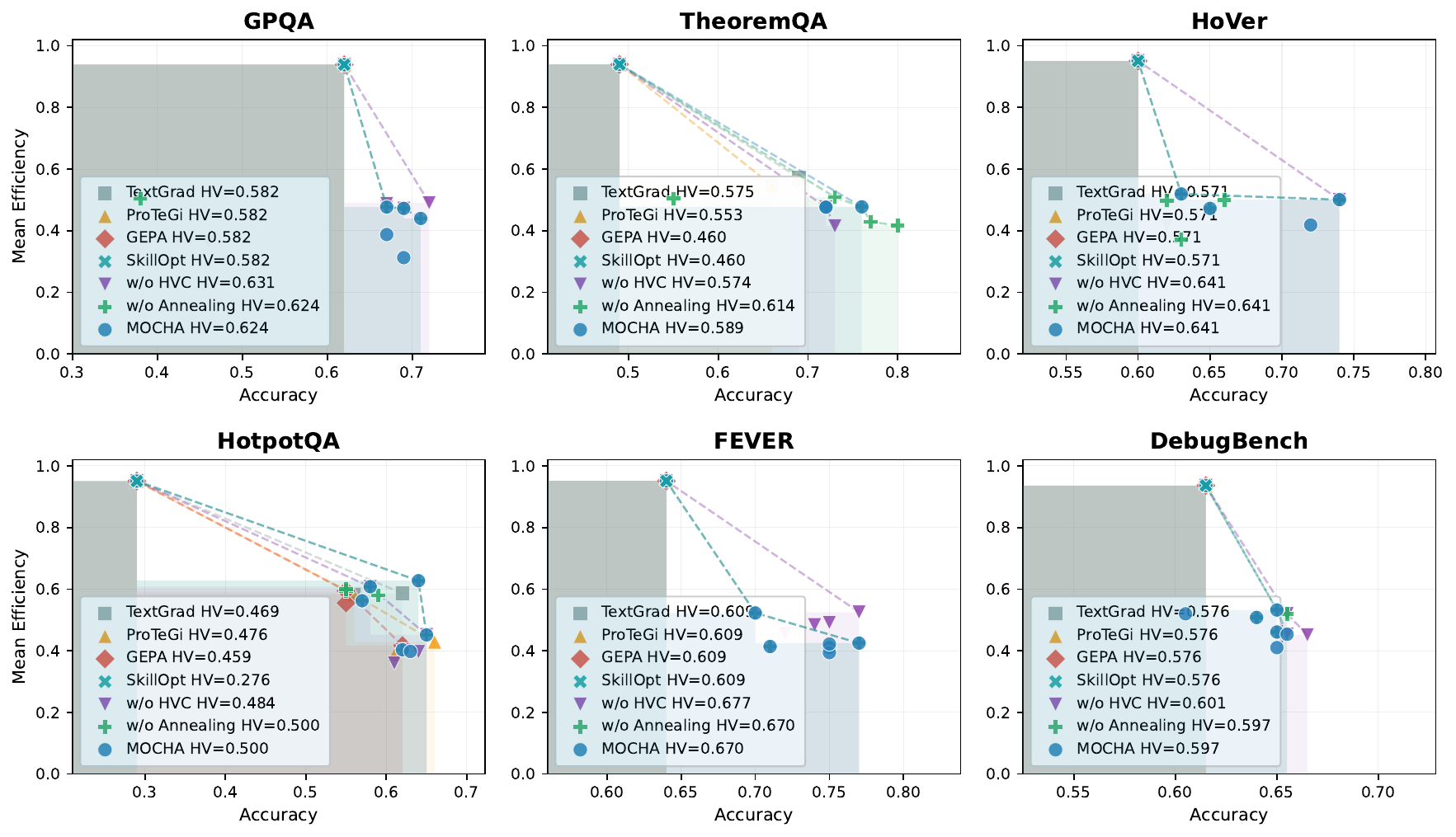}
\caption{\textbf{2D Pareto fronts} (accuracy $\times$ average efficiency) for all six skills. The pattern is consistent across all three views.}
\label{fig:pareto_2d_overall}
\end{figure*}

\Cref{fig:pareto_2d_body,fig:pareto_2d_desc,fig:pareto_2d_overall} show the fronts of all six skills in three views: accuracy against body efficiency, description efficiency, and their average. Without annealing (HVC throughout), MOCHA keeps the most candidates; without HVC (Chebyshev only), it pushes furthest on accuracy; full MOCHA balances both. The baselines mostly stay at a single point.

\subsection{Skill Evolution Trees}
\label{app:evolution_trees}

\begin{figure*}[h]
\centering
\includegraphics[width=\textwidth]{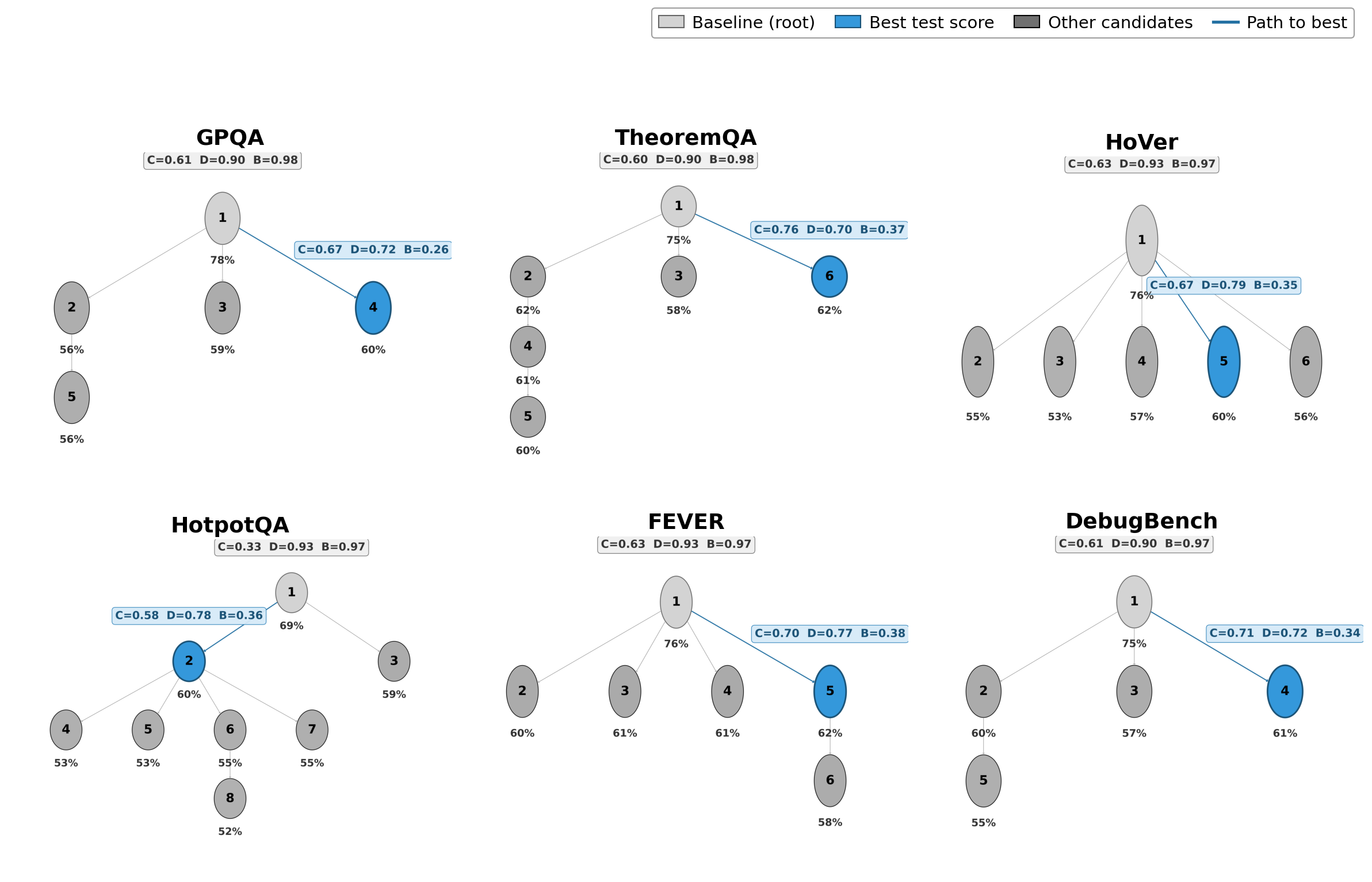}
\caption{\textbf{Skill evolution trees} for MOCHA across all six skills (shown from a random seed). Node labels show the candidate ID and test score; the blue node is the most accurate skill and blue edges trace its lineage. C/D/B at the root and the best node give accuracy and description/body efficiency.}
\label{fig:evolution_trees}
\end{figure*}

\Cref{fig:evolution_trees} shows MOCHA's evolution tree for every skill (shown from a random seed). Each node is a committed skill; edges trace rewrites. MOCHA explores several branches from the seed, and the best skills often emerge from non-obvious lineages rather than greedy refinement.

\subsection{Per-Task Qualitative Analysis}
\label{app:qualitative}

\Cref{fig:skill_compare} (main text) and \Cref{fig:skill_gpqa,fig:skill_theoremqa,fig:skill_hover,fig:skill_hotpotqa,fig:skill_debugbench} contrast each seed with a MOCHA skill.
The pattern is consistent. Where the baselines return the seed, MOCHA writes a procedure: GPQA gains a six-step protocol that tests every answer option and ends with an adversarial self-check; HoVer and FEVER gain claim decomposition and an all-or-nothing rule for \textsc{supports}; DebugBench gains a catalogue of bug patterns and an instruction to make minimal repairs. Where the baselines move, MOCHA's skills are more specific: TheoremQA names the governing theorem and fixes the numeric answer format, and HotpotQA enforces verbatim, minimal answers.

\begin{figure*}[t]
\centering
\begin{skillbox}[colback=red!3, colframe=red!50!black,
  fonttitle=\small, boxrule=0.5pt, left=3pt, right=3pt, top=2pt, bottom=2pt,
  arc=2pt]{\small\bfseries Seed Skill --- GPQA (Graduate STEM QA) \quad\scriptsize Accuracy: 0.59}
\footnotesize\ttfamily
\textcolor{gray}{---}\\
\textcolor{blue!70}{name:} gpqa\_solver \quad
\textcolor{blue!70}{compatibility:} dspy>=2.5, python>=3.10\\
\textcolor{blue!70}{metadata:} task: gpqa, modules: 1 \quad
\textcolor{blue!70}{allowed-tools:} []\\[1pt]
\textcolor{blue!70}{description:} Graduate-level STEM multiple-choice question answering
\hfill\scriptsize\textnormal{52/1,024 chars}\\[1pt]
\textcolor{gray}{---}\\[3pt]
\textcolor{gray}{\# execute.predict}\\
Given the fields `question', produce the fields `answer'.\\[6pt]
\textcolor{gray}{\itshape (no additional instructions)}\\[6pt]
\hfill\scriptsize\textnormal{57/5,000 chars}\\[4pt]
\textcolor{red!60!black}{\scriptsize\textnormal{$\blacktriangleright$ Returned unchanged by TextGrad, ProTeGi, and GEPA after 1000 rollouts}}
\end{skillbox}

\vspace{0.4em}

\begin{skillbox}[colback=green!3, colframe=green!50!black,
  fonttitle=\small, boxrule=0.5pt, left=3pt, right=3pt, top=2pt, bottom=2pt,
  arc=2pt]{\small\bfseries MOCHA-Optimized --- GPQA (Graduate STEM QA) \quad\scriptsize Accuracy: 0.71}
\footnotesize\ttfamily
\textcolor{gray}{---}\\
\textcolor{blue!70}{description:} Graduate-level STEM multiple-choice question answering requiring rigorous step-by-step scientific reasoning, deep domain expertise in physics, chemistry, biology, and mathematics, careful elimination of wrong answers through explicit testing of every option, adversarial self-checking for common pitfalls, and systematic verification before committing to the final answer letter.
\hfill\scriptsize\textnormal{$\sim$360/1,024 chars}\\[1pt]
\textcolor{gray}{---}\\[3pt]
\textcolor{gray}{\# execute.predict}\\
You are an expert-level STEM scientist.\\[2pt]
\textbf{Step 1: Deep Problem Parsing} --- Read question THREE times. 1st: identify domain. 2nd: extract values, conditions, reagents. 3rd: identify exactly what is asked. Note how answer choices differ from each other.\\[2pt]
\textbf{Step 2: Solve from First Principles} --- Work through fundamental laws/mechanisms. Write all equations explicitly.\\
~~$\bullet$ \textit{Chemistry (organic)}: Draw full structures incl.\ stereochemistry. Identify reactive site considering steric \& electronic effects at EVERY position. For named reactions (Wittig, Tebbe, Petasis, McMurry), recall exact mechanisms.\\
~~$\bullet$ \textit{Physics}: Set up equations, verify units at every step.\\
~~$\bullet$ \textit{Biology}: Map all crosses explicitly. Consider linkage, epistasis, penetrance.\\[2pt]
\textbf{Step 3: Test EVERY Answer Choice} --- For EACH option A--D, write verdict: ``A: fails because...'', ``B: matches because...''\\[2pt]
\textbf{Step 4: Adversarial Self-Check (MANDATORY)} --- Assume first answer is WRONG. Actively seek errors:\\
~~$\bullet$ Steric hindrance: correctly identify LEAST hindered site?\\
~~$\bullet$ Reaction selectivity: correct reacting position?\\
~~$\bullet$ Sign/direction errors, symmetry, meso compounds, approximation validity\\
~~$\bullet$ Named reaction specifics: Tebbe converts esters to enol ethers, Simmons-Smith gives cyclopropanation\\[2pt]
\textbf{Step 5: Compare Top Two Candidates} --- Head-to-head on single distinguishing factor.\\[2pt]
\textbf{Step 6: Commit} --- Single letter: A, B, C, or D. Favor standard mechanism over exotic.\\[2pt]
\textbf{Key Error Patterns:} (1) Steric effects in enolization: t-BuOK favors less substituted kinetic enolate. (2) Pathway sequences: trace causally. (3) Organometallic specificity: Cp$_2$TiMeCl vs Cp$_2$TiMe$_2$. (4) Don't conflate similar reactions.\\[2pt]
\hfill\scriptsize\textnormal{$\sim$3,418/5,000 chars}
\end{skillbox}
\caption{\textbf{GPQA example.} Seed (red) versus MOCHA (green): six-step scientific reasoning with adversarial checks. Illustrative skill with the highest test accuracy among committed skills; test accuracy $.59\to.71$.}
\label{fig:skill_gpqa}
\end{figure*}

\begin{figure*}[t]
\centering
\begin{skillbox}[colback=red!3, colframe=red!50!black,
  fonttitle=\small, boxrule=0.5pt, left=3pt, right=3pt, top=2pt, bottom=2pt,
  arc=2pt]{\small\bfseries Seed Skill --- TheoremQA (Mathematical Reasoning) \quad\scriptsize Accuracy: 0.53}
\footnotesize\ttfamily
\textcolor{gray}{---}\\
\textcolor{blue!70}{name:} theoremqa\_solver \quad
\textcolor{blue!70}{compatibility:} dspy>=2.5, python>=3.10\\
\textcolor{blue!70}{metadata:} task: theoremqa, modules: 1 \quad
\textcolor{blue!70}{allowed-tools:} []\\[1pt]
\textcolor{blue!70}{description:} Theorem-based QA across math, physics, and CS domains
\hfill\scriptsize\textnormal{49/1,024 chars}\\[1pt]
\textcolor{gray}{---}\\[3pt]
\textcolor{gray}{\# execute.predict}\\
Given the fields `problem', produce the fields `answer'.\\[6pt]
\textcolor{gray}{\itshape (no additional instructions)}\\[6pt]
\hfill\scriptsize\textnormal{57/5,000 chars}\\[4pt]
\textcolor{red!60!black}{\scriptsize\textnormal{$\blacktriangleright$ Partially optimized by baselines but with verbose, loosely structured output}}
\end{skillbox}

\vspace{0.4em}

\begin{skillbox}[colback=green!3, colframe=green!50!black,
  fonttitle=\small, boxrule=0.5pt, left=3pt, right=3pt, top=2pt, bottom=2pt,
  arc=2pt]{\small\bfseries MOCHA-Optimized --- TheoremQA (Mathematical Reasoning) \quad\scriptsize Accuracy: 0.82}
\footnotesize\ttfamily
\textcolor{gray}{---}\\
\textcolor{blue!70}{description:} Solves theorem-based questions across math, physics, finance, EE, and CS by carefully applying relevant theorems, tracking signs/units, and returning precisely formatted answers as decimal numbers, booleans, or lists.
\hfill\scriptsize\textnormal{$\sim$215/1,024 chars}\\[1pt]
\textcolor{gray}{---}\\[3pt]
\textcolor{gray}{\# execute.predict}\\
You are an expert problem solver across mathematics, physics, EE, CS, and finance.\\[2pt]
\textbf{Reasoning Protocol:}\\
\textbf{1. Identify the domain and relevant theorem(s)}: State which theorem, formula, or principle applies.\\
\textbf{2. Define all variables and given quantities explicitly}: Write out every given value with correct signs and units.\\
\textbf{3. Apply the theorem step by step}: Show each algebraic/logical step. Double-check:\\
~~$\bullet$ \textbf{Signs}: Pay extreme attention to negative signs. Never drop them.\\
~~$\bullet$ \textbf{Powers of 10}: Verify exponent arithmetic carefully.\\
~~$\bullet$ \textbf{Units}: Track throughout. Convert as needed but CHECK expected units.\\[2pt]
\textbf{4. Domain-specific rules:}\\
~~$\bullet$ \textit{Resistance with geometry}: Axial: $R = \rho L / (\pi(R_o^2 - R_i^2))$. Radial: $R = (\rho/2\pi L) \ln(R_o/R_i)$.\\
~~$\bullet$ \textit{Stopping times}: $T$ is stopping time iff $\{T \leq t\} \in \mathcal{F}_t$. Sum of non-negative stopping times IS a stopping time.\\
~~$\bullet$ \textit{Iteration methods}: For Aitken's $\Delta^2$, count iterations of the ACCELERATED method only.\\[2pt]
\textbf{CRITICAL Formatting Rules:}\\
$\bullet$ If multiple sub-parts, return ONLY the answer to the LAST part asked.\\
$\bullet$ Return ONLY the final answer---no explanations, no units, no labels.\\
$\bullet$ Convert ALL symbolic/fractional answers to decimals: \texttt{4/9} $\to$ \texttt{0.4444}, \texttt{sqrt(2)} $\to$ \texttt{1.4142}.\\
$\bullet$ Floats ALWAYS include decimal point: \texttt{3.0} not \texttt{3}.\\
$\bullet$ Round to 4 decimal places unless exact.\\[2pt]
\hfill\scriptsize\textnormal{$\sim$2,517/5,000 chars}
\end{skillbox}
\caption{\textbf{TheoremQA example.} MOCHA adds theorem identification, sign/unit checks, and output rules. Illustrative skill with the highest test accuracy among committed skills; test accuracy $.53\to.82$.}
\label{fig:skill_theoremqa}
\end{figure*}

\begin{figure*}[t]
\centering
\begin{skillbox}[colback=red!3, colframe=red!50!black,
  fonttitle=\small, boxrule=0.5pt, left=3pt, right=3pt, top=2pt, bottom=2pt,
  arc=2pt]{\small\bfseries Seed Skill --- HoVer (Multi-hop Claim Verification) \quad\scriptsize Accuracy: 0.62}
\footnotesize\ttfamily
\textcolor{gray}{---}\\
\textcolor{blue!70}{name:} hover\_verification \quad
\textcolor{blue!70}{compatibility:} dspy>=2.5, python>=3.10\\
\textcolor{blue!70}{metadata:} task: hover\_singlecall, modules: 1 \quad
\textcolor{blue!70}{allowed-tools:} [retriever]\\[1pt]
\textcolor{blue!70}{description:} Claim verification with retrieval tool
\hfill\scriptsize\textnormal{38/1,024 chars}\\[1pt]
\textcolor{gray}{---}\\[3pt]
\textcolor{gray}{\# execute.predict}\\
Given the fields `claim', `evidence', produce the fields `verdict'.\\[6pt]
\textcolor{gray}{\itshape (no additional instructions)}\\[6pt]
\hfill\scriptsize\textnormal{67/5,000 chars}\\[4pt]
\textcolor{red!60!black}{\scriptsize\textnormal{$\blacktriangleright$ Returned unchanged by TextGrad, ProTeGi, and GEPA after 1000 rollouts}}
\end{skillbox}

\vspace{0.4em}

\begin{skillbox}[colback=green!3, colframe=green!50!black,
  fonttitle=\small, boxrule=0.5pt, left=3pt, right=3pt, top=2pt, bottom=2pt,
  arc=2pt]{\small\bfseries MOCHA-Optimized --- HoVer (Multi-hop Claim Verification) \quad\scriptsize Accuracy: 0.67}
\footnotesize\ttfamily
\textcolor{gray}{---}\\
\textcolor{blue!70}{description:} Multi-hop claim verification against retrieved evidence passages. Decomposes complex claims into sub-claims, retrieves and chains evidence, and produces a binary SUPPORTED or NOT\_SUPPORTED verdict.
\hfill\scriptsize\textnormal{$\sim$198/1,024 chars}\\[1pt]
\textcolor{gray}{---}\\[3pt]
\textcolor{gray}{\# execute.predict}\\
You are an expert fact-checker performing multi-hop claim verification.\\[2pt]
\textbf{Critical Rules for Verdict:}\\
\textbf{1.} Binary output only: exactly \texttt{SUPPORTED} or \texttt{NOT\_SUPPORTED}. Never ``PARTIALLY SUPPORTED'' or any other value.\\
\textbf{2.} Default toward SUPPORTED when evidence is consistent. Do NOT require every detail to be explicitly stated---implicit support and reasonable inference count.\\
\textbf{3.} Only NOT\_SUPPORTED when evidence \textit{actively contradicts} the claim.\\[2pt]
\textbf{Reasoning Strategy:}\\
\textbf{Step 1}: Decompose claim into atomic sub-claims.\\
\textbf{Step 2}: Map evidence to sub-claims. Note direct vs.\ inferential support.\\
\textbf{Step 3}: Use retriever tool to fill gaps with targeted queries.\\
\textbf{Step 4}: Chain reasoning across passages. Follow entity links completely.\\
\textbf{Step 5}: Check for explicit contradictions.\\
\textbf{Step 6}: For comparative/temporal claims, extract specific quantities and compare directly.\\
\textbf{Step 7}: Produce verdict.\\[2pt]
\textbf{Common Mistakes to Avoid:}\\
$\bullet$ Do NOT default to NOT\_SUPPORTED because evidence is incomplete---retrieve more.\\
$\bullet$ Do NOT require verbatim confirmation. Logical inference counts.\\
$\bullet$ Do NOT conflate ``not mentioned'' with ``contradicted.''\\
$\bullet$ Absence of evidence is not evidence of absence---retrieve more evidence.\\[2pt]
\hfill\scriptsize\textnormal{$\sim$2,141/5,000 chars}
\end{skillbox}
\caption{\textbf{HoVer example.} MOCHA adds retrieval-backed verification and distinguishes missing evidence from contradiction. Illustrative skill with the highest test accuracy among committed skills; test accuracy $.62\to.67$.}
\label{fig:skill_hover}
\end{figure*}

\begin{figure*}[t]
\centering
\begin{skillbox}[colback=red!3, colframe=red!50!black,
  fonttitle=\small, boxrule=0.5pt, left=3pt, right=3pt, top=2pt, bottom=2pt,
  arc=2pt]{\small\bfseries Seed Skill --- HotpotQA (Multi-hop QA) \quad\scriptsize Accuracy: 0.34}
\footnotesize\ttfamily
\textcolor{gray}{---}\\
\textcolor{blue!70}{name:} hotpotqa\_qa \quad
\textcolor{blue!70}{compatibility:} dspy>=2.5, python>=3.10\\
\textcolor{blue!70}{metadata:} task: hotpotqa\_singlecall, modules: 1 \quad
\textcolor{blue!70}{allowed-tools:} [retriever]\\[1pt]
\textcolor{blue!70}{description:} Question answering with retrieval tool
\hfill\scriptsize\textnormal{38/1,024 chars}\\[1pt]
\textcolor{gray}{---}\\[3pt]
\textcolor{gray}{\# execute.predict}\\
Given the fields `question', `evidence', produce the fields `answer'.\\[6pt]
\textcolor{gray}{\itshape (no additional instructions)}\\[6pt]
\hfill\scriptsize\textnormal{67/5,000 chars}\\[4pt]
\textcolor{red!60!black}{\scriptsize\textnormal{$\blacktriangleright$ Partially optimized by all baselines}}
\end{skillbox}

\vspace{0.4em}

\begin{skillbox}[colback=green!3, colframe=green!50!black,
  fonttitle=\small, boxrule=0.5pt, left=3pt, right=3pt, top=2pt, bottom=2pt,
  arc=2pt]{\small\bfseries MOCHA-Optimized --- HotpotQA (Multi-hop QA) \quad\scriptsize Accuracy: 0.66}
\footnotesize\ttfamily
\textcolor{gray}{---}\\
\textcolor{blue!70}{description:} Multi-hop question answering over retrieved evidence passages, producing short exact answers extracted verbatim from evidence text
\hfill\scriptsize\textnormal{$\sim$131/1,024 chars}\\[1pt]
\textcolor{gray}{---}\\[3pt]
\textcolor{gray}{\# execute.predict}\\
Given the fields `question', `evidence', produce the fields `answer'.\\[2pt]
\textbf{Instructions for Reasoning:}\\
\textbf{1. Decompose}: Break question into sub-questions. Identify what entity/fact each hop requires.\\
\textbf{2. Extract}: Read every evidence piece. Extract all names (full formal names), dates, nicknames, roles, locations---even from parenthetical remarks.\\
\textbf{3. Retrieve}: If evidence is insufficient, call retriever with targeted queries. Do NOT give up.\\
\textbf{4. Chain}: Connect facts across passages. Entity A in passage 1 $\to$ Entity B in passage 2.\\
\textbf{5. Synthesize}: Determine final answer.\\[2pt]
\textbf{Critical Rules for Answer Field:}\\
$\bullet$ Short exact phrase---name, date, number, place, or brief noun phrase.\\
$\bullet$ \textbf{EXACT form from evidence}: ``Jerral Wayne Jones Sr.'' NOT ``Jerry Jones''. ``Dayton, Ohio'' NOT ``Dayton''.\\
$\bullet$ Copy verbatim whenever possible. Preserve location qualifiers.\\
$\bullet$ Use the most complete, formal name version found in evidence.\\
$\bullet$ Do NOT add explanations, parenthetical remarks, or full sentences.\\
$\bullet$ Strip ``the answer is'' but keep geographic qualifiers, suffixes (Sr., Jr., III).\\
$\bullet$ Never say ``cannot be determined''---use the retriever tool first.\\[2pt]
\textbf{Correct formatting examples:}\\
~~Good: \texttt{Dayton, Ohio} $\vert$ Bad: \texttt{Dayton}\\
~~Good: \texttt{Jerral Wayne Jones Sr.} $\vert$ Bad: \texttt{Jerry Jones}\\
~~Good: \texttt{Tony Yoka} $\vert$ Bad: \texttt{Tony Yoka, a French boxer}\\
~~Good: \texttt{1953} $\vert$ Bad: \texttt{The answer is 1953}\\[2pt]
\hfill\scriptsize\textnormal{$\sim$2,593/5,000 chars}
\end{skillbox}
\caption{\textbf{HotpotQA example.} MOCHA adds multi-hop reasoning and verbatim answer extraction with formatting examples. Illustrative skill with the highest test accuracy among committed skills; test accuracy $.34\to.66$.}
\label{fig:skill_hotpotqa}
\end{figure*}

\begin{figure*}[t]
\centering
\begin{skillbox}[colback=red!3, colframe=red!50!black,
  fonttitle=\small, boxrule=0.5pt, left=3pt, right=3pt, top=2pt, bottom=2pt,
  arc=2pt]{\small\bfseries Seed Skill --- DebugBench (Code Debugging) \quad\scriptsize Accuracy: 0.62}
\footnotesize\ttfamily
\textcolor{gray}{---}\\
\textcolor{blue!70}{name:} debugbench\_fixer \quad
\textcolor{blue!70}{compatibility:} dspy>=2.5, python>=3.10\\
\textcolor{blue!70}{metadata:} task: debugbench\_singlecall, modules: 1 \quad
\textcolor{blue!70}{allowed-tools:} []\\[1pt]
\textcolor{blue!70}{description:} Single-call code debugging (buggy\_code -> fixed\_code)
\hfill\scriptsize\textnormal{48/1,024 chars}\\[1pt]
\textcolor{gray}{---}\\[3pt]
\textcolor{gray}{\# execute.predict}\\
Given the fields `buggy\_code', `bug\_type', produce the fields `fixed\_code'.\\[6pt]
\textcolor{gray}{\itshape (no additional instructions)}\\[6pt]
\hfill\scriptsize\textnormal{75/5,000 chars}\\[4pt]
\textcolor{red!60!black}{\scriptsize\textnormal{$\blacktriangleright$ Returned unchanged by TextGrad, ProTeGi, and GEPA after 1000 rollouts}}
\end{skillbox}

\vspace{0.4em}

\begin{skillbox}[colback=green!3, colframe=green!50!black,
  fonttitle=\small, boxrule=0.5pt, left=3pt, right=3pt, top=2pt, bottom=2pt,
  arc=2pt]{\small\bfseries MOCHA-Optimized --- DebugBench (Code Debugging) \quad\scriptsize Accuracy: 0.73}
\footnotesize\ttfamily
\textcolor{gray}{---}\\
\textcolor{blue!70}{description:} Fix bugs in code based on the specified bug\_type classification. For single-category bugs, make the minimal change needed. For multiple errors, fix all bugs found. Do not refactor or improve beyond what the bug\_type requires.
\hfill\scriptsize\textnormal{$\sim$225/1,024 chars}\\[1pt]
\textcolor{gray}{---}\\[3pt]
\textcolor{gray}{\# execute.predict}\\
You are a precise code debugger. You receive buggy\_code and a bug\_type classification.\\[2pt]
\textbf{Critical Rules:}\\
\textbf{1. Understand bug\_type first}---it determines fixing strategy:\\
~~$\bullet$ \texttt{reference error}: Wrong variable/function/method name. Fix ONLY the incorrect reference(s).\\
~~$\bullet$ \texttt{syntax error}: Missing colon, semicolon, bracket, wrong operator syntax. Fix ONLY syntax.\\
~~$\bullet$ \texttt{logic error}: Off-by-one, wrong comparison, wrong return, wrong condition. Fix ONLY logic.\\
~~$\bullet$ \texttt{type error}: Wrong type usage or conversion.\\
~~$\bullet$ \texttt{multiple error}: 2--4 bugs across categories. Find and fix ALL.\\[2pt]
\textbf{2. Conservative fixing principle}: When uncertain, do NOT change. A wrong fix is worse than a missing fix.\\[2pt]
\textbf{3. Reproduce the rest EXACTLY}---preserve all indentation, spacing, comments, structure.\\[2pt]
\textbf{Reasoning Process:}\\
\textbf{Step 1}: Read \texttt{bug\_type}. Single-category or multiple?\\
\textbf{Step 2}: Understand algorithm PURPOSE before making changes.\\
\textbf{Step 3}: For each bug, state: (a) exact line, (b) what is wrong, (c) fix, (d) why it is definitely a bug.\\
\textbf{Step 4}: For \texttt{multiple error}---count bugs. Typical: 2--4. If only 1, look harder. If $>$5, reconsider.\\
\textbf{Step 5}: For reference errors---replace with correct name that EXISTS in scope.\\
\textbf{Step 6}: Apply fixes. Double-check each is minimal and correct.\\
\textbf{Step 7}: Re-read output. Confirm only bug\_type-matching changes were made.\\[2pt]
\textbf{Common Pitfalls:}\\
$\bullet$ Do NOT change loop bounds unless certain they cause error.\\
$\bullet$ Do NOT treat \texttt{multiple error} as license to rewrite---each fix is a small, targeted change.\\
$\bullet$ When variable exists elsewhere in code, prefer that exact spelling as the fix.\\
$\bullet$ For C++/Java: common bugs include missing semicolons, \texttt{=} vs \texttt{==}, wrong variable names.\\[2pt]
\hfill\scriptsize\textnormal{$\sim$2,315/5,000 chars}
\end{skillbox}
\caption{\textbf{DebugBench example.} MOCHA replaces the seed template with bug-type-specific checks and conservative, minimal repairs.}
\label{fig:skill_debugbench}
\end{figure*}

\subsection{Per-Task Ablation Detail}
\label{app:ablation_detail}

\Cref{tab:ablation_detail} gives per-task accuracy for each MOCHA variant and the baselines, complementing \Cref{tab:ablation}.

\begin{table*}[t]
\centering
\caption{\textbf{Per-task ablations} for Claude Haiku~4.5 (test accuracy, mean$\pm$std over five seeds). \textbf{Bold}: best per task.}
\label{tab:ablation_detail}
\footnotesize
\setlength{\tabcolsep}{3pt}
\begin{tabular}{@{}lcccccc@{}}
\toprule
\textbf{Skill} & \textbf{TextGrad} & \textbf{ProTeGi} & \textbf{GEPA} & \textbf{w/o HVC} & \textbf{w/o Ann.} & \textbf{MOCHA} \\
\midrule
GPQA & .592\scriptsize$\pm$.025 & .592\scriptsize$\pm$.025 & .592\scriptsize$\pm$.025 & \textbf{.638\scriptsize$\pm$.040} & .636\scriptsize$\pm$.056 & .636\scriptsize$\pm$.056 \\
TheoremQA & .672\scriptsize$\pm$.087 & .690\scriptsize$\pm$.105 & .656\scriptsize$\pm$.129 & \textbf{.776\scriptsize$\pm$.031} & .770\scriptsize$\pm$.039 & .762\scriptsize$\pm$.046 \\
HoVer & .618\scriptsize$\pm$.015 & .618\scriptsize$\pm$.015 & .618\scriptsize$\pm$.015 & .658\scriptsize$\pm$.047 & .614\scriptsize$\pm$.109 & \textbf{.660\scriptsize$\pm$.043} \\
HotpotQA & .592\scriptsize$\pm$.030 & \textbf{.622\scriptsize$\pm$.041} & .602\scriptsize$\pm$.029 & .602\scriptsize$\pm$.040 & .608\scriptsize$\pm$.036 & .600\scriptsize$\pm$.034 \\
FEVER & .632\scriptsize$\pm$.032 & .632\scriptsize$\pm$.032 & .632\scriptsize$\pm$.032 & \textbf{.754\scriptsize$\pm$.023} & .722\scriptsize$\pm$.026 & .726\scriptsize$\pm$.027 \\
DebugBench & .615\scriptsize$\pm$.007 & .615\scriptsize$\pm$.007 & .615\scriptsize$\pm$.007 & \textbf{.692\scriptsize$\pm$.027} & .678\scriptsize$\pm$.014 & .666\scriptsize$\pm$.039 \\
\midrule
\textbf{Mean} & .620 & .628 & .619 & \textbf{.687} & .671 & .675 \\
\bottomrule
\end{tabular}

\end{table*}

All MOCHA variants clearly outperform the baselines on GPQA, FEVER, and DebugBench, where the baselines return the seed unchanged. Removing HVC gives the highest accuracy on four of six tasks, consistent with its place at the exploitation end of the spectrum; full MOCHA is best on HoVer, where annealing prevents the premature convergence of the explore-only variant.

\subsection{Ablation Heatmap}
\label{app:ablation}

\begin{figure}[h]
\centering
\includegraphics[width=\linewidth]{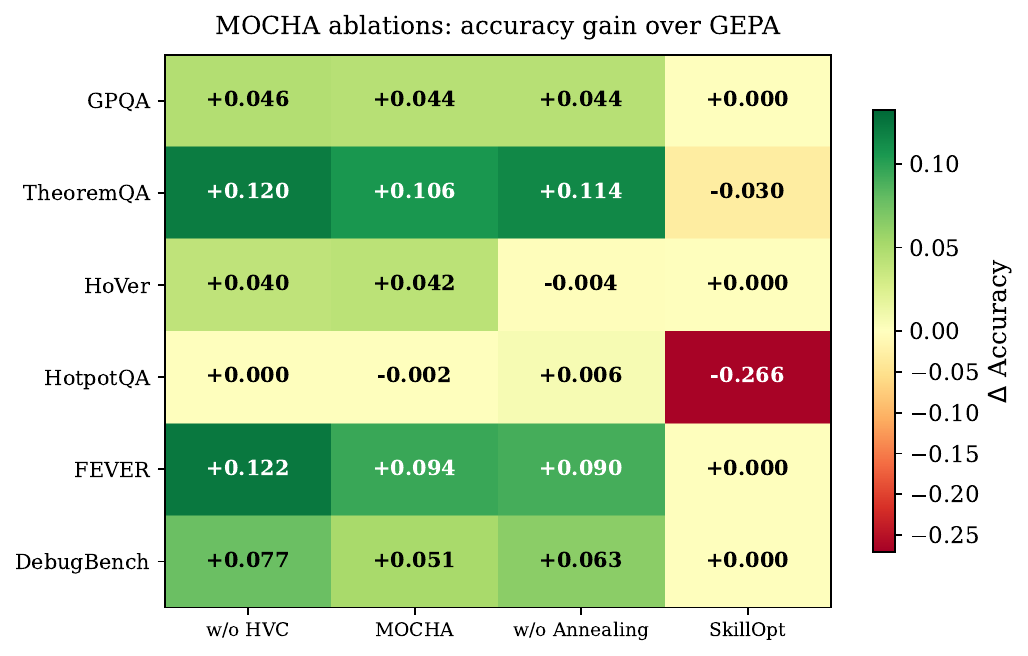}
\caption{\textbf{Ablation heatmap}: accuracy gain over GEPA for each MOCHA variant and for SkillOpt, per skill.}
\label{fig:ablation_app}
\end{figure}

\Cref{fig:ablation_app} shows the per-skill accuracy of \Cref{tab:ablation_detail} as gains over GEPA, and adds SkillOpt. All MOCHA variants gain most where GEPA stalls (GPQA, FEVER, DebugBench) and on TheoremQA; on HotpotQA, whose weak seed every optimizer improves, they are within $.006$ of GEPA. SkillOpt, which returns the seed in nearly every run, gains over GEPA on no skill and falls well below it on HotpotQA, the one skill where GEPA moves far beyond the seed.

\section{Broader Impact}
\label{app:broader_impact}

MOCHA is a general-purpose optimizer for structured natural-language skill specifications. Its primary application is improving the accuracy and token efficiency of LLM agent skills. We do not foresee direct negative societal impacts beyond those inherent to LLM deployment: optimized skills could in principle be applied to harmful tasks, a risk shared with all prompt optimization methods. MOCHA does not introduce new capabilities; it refines existing skill definitions through API-level access only.

\section{Compute Resources}
\label{app:compute}

Each run (one method, task, and seed) consumes 1000 rollouts and takes about two to four hours of wall-clock time through API calls. All computation used cloud APIs; no GPUs were required.

\section{Reproducibility and Code Availability}
\label{app:reproducibility}

MOCHA requires only API access to an LLM provider. \Cref{app:pseudocode} gives Python-like pseudocode of the full algorithm, and \Cref{app:agents} an \texttt{AGENTS.md} guide for re-implementing it and reproducing our experiments. All hyperparameters are given in \Cref{app:implementation}; seeds (0, 42, 123, 456, 786) and data splits (100 train / 100 validation / 100 test) are fixed across experiments.

\textbf{Dataset licenses.} GPQA~\citep{rein2023gpqa}: CC-BY-4.0. TheoremQA~\citep{chen2023theoremqa}: MIT. HoVer~\citep{jiang2020hover}: CC-BY-SA-4.0. HotpotQA~\citep{yang2018hotpotqa}: CC-BY-SA-4.0. FEVER~\citep{thorne2018fever}: CC-BY-SA-3.0. DebugBench~\citep{tian2024debugbench}: Apache-2.0. All datasets are used in accordance with their licenses for research purposes.

\end{document}